\documentclass[10pt,twocolumn,letterpaper]{article}

\usepackage{cvpr}
\usepackage{times}
\usepackage{epsfig}
\usepackage{graphicx}
\usepackage{amsmath}
\usepackage{amssymb}

\usepackage[tableposition=above]{caption}
\usepackage{subcaption}
\usepackage{bbm}
\usepackage{bm}
\usepackage{xspace}
\usepackage[utf8]{inputenc} % allow utf-8 input
\usepackage{url}            % simple URL typesetting
\usepackage{booktabs}       % professional-quality tables
\usepackage{amsfonts}       % blackboard math symbols
\usepackage{nicefrac}       % compact symbols for 1/2, etc.
\usepackage{microtype}      % microtypography
\usepackage{color}
\usepackage{algorithm}
\usepackage{algorithmic}
\usepackage{array}
\usepackage{multirow}
\usepackage[para,online,flushleft]{threeparttable}
\usepackage{float}
\usepackage{enumitem}
\usepackage{makecell}
\usepackage{gensymb}         % for degree
\usepackage{mathtools}

\makeatletter
\DeclareRobustCommand\onedot{\futurelet\@let@token\@onedot}
\def\@onedot{\ifx\@let@token.\else.\null\fi\xspace}

\def\eg{\emph{e.g}\onedot} 
\def\ie{\emph{i.e}\onedot}

\makeatother

\DeclareMathOperator{\arctantwo}{arctan2}

\newcommand{\threesixty}{360$\degree$~}

\usepackage{xcolor}
\usepackage{ifthen}
\usepackage{textcomp}
\definecolor{red}{rgb}{0.9,0.1,0}
\definecolor{slateblue}{rgb}{0.7,0.35,0.9}
\definecolor{green}{rgb}{0, 0.4, 0}
\definecolor{brown}{rgb}{0.3, 0.2, 0}
\definecolor{mahogany}{rgb}{0.75, 0.25, 0.0}
\definecolor{purple}{rgb}{0.3, 0, 0.3}
\definecolor{darkgreen}{rgb}{0, 0.4, 0}
\definecolor{frenchblue}{rgb}{0.0, 0.45, 0.73}
\definecolor{blue}{rgb}{0.0, 0.0, 1.0}
\definecolor{goldenrod}{rgb}{0.65, 0.45, 0.03}
\definecolor{gray}{rgb}{0.5,0.5,0.5}

\newboolean{revising}
\setboolean{revising}{true}
\ifthenelse{\boolean{revising}}
{
    \newcommand{\ignore}[1]{}
    
} {
    \newcommand{\ignore}[1]{}
    
}

\makeatletter
\renewcommand{\paragraph}{%
  \@startsection{paragraph}{4}%
  {\z@}{0.5\baselineskip \@plus 0ex \@minus 0ex}{-1em}%
  {\normalfont\normalsize\bfseries}%
}
\makeatother

\usepackage[breaklinks=true,bookmarks=false,colorlinks]{hyperref}

\cvprfinalcopy % *** Uncomment this line for the final submission

\def\cvprPaperID{****} % *** Enter the CVPR Paper ID here

\makeatletter
\let\@fnsymbol\@arabic
\makeatother

\begin{document}

%%%%%%%%% TITLE
\title{Indoor Panorama Planar 3D Reconstruction via Divide and Conquer}

\author{
Cheng Sun\thanks{National Tsing Hua University}~$^{,}$\thanks{ASUS AICS Department} \\ \href{mailto:chengsun@gapp.nthu.edu.tw}{\small\tt \textcolor{black}{chengsun@gapp.nthu.edu.tw}}
\and
Chi-Wei Hsiao\footnotemark[1] \\ \href{mailto:chiweihsiao@gapp.nthu.edu.tw}{\small\tt \textcolor{black}{chiweihsiao@gapp.nthu.edu.tw}}
\and
Ning-Hsu Wang\footnotemark[1] \\ \href{mailto:albert100121@gapp.nthu.edu.tw}{\small\tt \textcolor{black}{albert100121@gapp.nthu.edu.tw}}
\and
Min Sun\footnotemark[1]~$^{,}$\thanks{Joint Research Center for AI Technology and All Vista Healthcare} \\ \href{mailto:sunmin@ee.nthu.edu.tw}{\small\tt \textcolor{black}{sunmin@ee.nthu.edu.tw}}
\and
Hwann-Tzong Chen\footnotemark[1]~$^{,}$\thanks{Aeolus Robotics} \\ \href{mailto:htchen@cs.nthu.edu.tw}{\small\tt \textcolor{black}{htchen@cs.nthu.edu.tw}}
}

\maketitle
\thispagestyle{empty}

%%%%%%%%% ABSTRACT
\begin{abstract}
Indoor panorama typically consists of human-made structures parallel or perpendicular to gravity.
We leverage this phenomenon to approximate the scene in a 360-degree image with (H)orizontal-planes and (V)ertical-planes.
To this end, we propose an effective divide-and-conquer strategy that divides pixels based on their plane orientation estimation; then, the succeeding instance segmentation module conquers the task of planes clustering more easily in each plane orientation group.
Besides, parameters of V-planes depend on camera yaw rotation, but translation-invariant CNNs are less aware of the yaw change.
We thus propose a yaw-invariant V-planar reparameterization for CNNs to learn.
We create a benchmark for indoor panorama planar reconstruction by extending existing 360 depth datasets with ground truth H\&V-planes (referred to as ``PanoH\&V'' dataset) and adopt state-of-the-art planar reconstruction methods to predict H\&V-planes as our baselines.
Our method outperforms the baselines by a large margin on the proposed dataset.
Code is available at \url{https://github.com/sunset1995/PanoPlane360}.
\end{abstract}

\vspace{-2em}

%%%%%%%%% Introduction start
\section{Introduction}

Reconstructing planar surfaces from single-view images have many applications such as interior modeling, AR/VR, robot navigation, and scene understanding.
Generally, an indoor scene consisting of human-made structures can be approximated by a small set of dominant planes, making planar reconstruction suitable for 3D indoor modeling.

Recent works on planar reconstruction~\cite{LiuKGFK19,LiuYCYF18,NewellHD17} employ state-of-the-art instance segmentation methods and achieve promising results.
However, these works are mostly trained on the planar datasets derived from ScanNet~\cite{DaiCSHFN17} or NYUv2~\cite{SilbermanHKF12} with a small field-of-view (FoV).
Data of this kind require multiple images to reconstruct entire scenes, which would cost more computational time and resources.
As \threesixty devices get popularized, the amount of \threesixty data has significantly increased, with many panorama datasets~\cite{ChangDFHNSSZZ17, ArmeniSZS17, WangSTCS20} being released to facilitate learning-based methods.
By taking input in \threesixty format, 3D reconstruction of an entire scene can be done with only one snapshot.
Considering the benefit of real-world \threesixty data in planar reconstruction and the gap of existing literature in this research field, we believe the planar reconstruction task from panorama imagery is worthy of investigation.

\begin{figure}
    \centering
    \includegraphics[width=\linewidth]{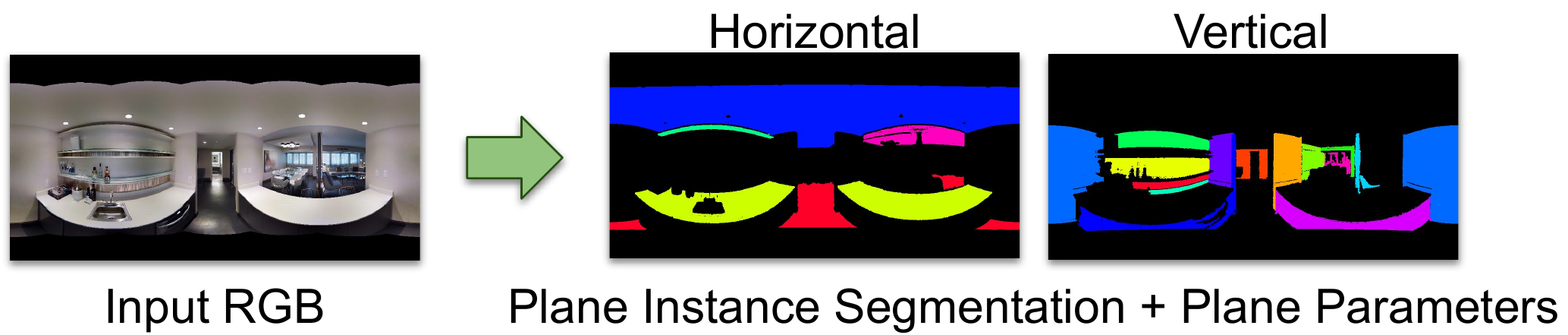}
    \caption{
        Planar surfaces of human-made structures are mostly horizontal or vertical with respect to the gravity direction.
        Given an RGB panorama, we propose to model a 3D scene as horizontal or vertical planes (the \emph{H\&V-planes}).
    }
    \label{fig:teaser}
    \vspace{-1em}
\end{figure}

In this work, we construct the first real-world indoor \threesixty H\&V-plane dataset (PanoH\&V dataset), where we simplify the planar reconstruction task by focusing on horizontal and vertical planes (illustrated in Fig.~\ref{fig:teaser}).
To this end, we extend existing large-scale \threesixty datasets by fitting the provided depth modality with horizontal and vertical planes.
The H\&V-planes are similar to the concept of the Manhattan world~\cite{CoughlanY00} and the Atlanta world~\cite{SchindlerD04}, but we only constrain the extracted planes to be vertical or horizontal without assuming other inter-plane relationship.
With the extended modality from large-scale H\&V-planes, we train two current state-of-the-art planar reconstruction methods, PlaneRCNN~\cite{LiuKGFK19} and PlaneAE~\cite{YuZLZG19}, and report their performance to serve as the strong baselines in our benchmark.

We find existing planar reconstruction methods suboptimal when applied to the presented PanoH\&V dataset.
First, existing methods employ instance segmentation to detect planes from visual cue with less consideration about the estimated geometry.
In practice, some planes are easier to differentiate through geometry instead of visual appearance (\eg, walls with similar appearance).
Second, plane parameters depend on camera poses, but \threesixty camera yaw rotations are left-right circular shifts on equirectangular images and hard for the translation-invariant CNNs to observe.
In contrast to the existing works, our method addresses the above issues appropriately:
{\it i)} We use a divide-and-conquer strategy. 
The proposed \emph{surface orientation grouping} distinguishes pixels of different plane orientations, so the succeeding instance segmentation module applied to each group can focus on a simpler subproblem.
{\it ii)} We then propose a residual form \emph{yaw-invariant parameterization} for V-planar geometry such that it is independent of the camera yaw rotation.
We show that the yaw-invariant parameterization brings significant improvements in V-plane orientations estimation, which also benefits other methods.

We summarize the contribution of this work in two aspects.
In terms of \textbf{technical contribution}, the proposed method consists of \textit{i}) a divide-and-conquer strategy for the task of plane instance segmentation, which exploits the estimated plane orientations to divide the task into multiple simpler subproblems; \textit{ii}) a yaw-invariant vertical plane parameterization addressing the \threesixty yaw ambiguity, which can also boost other existing methods.
In terms of \textbf{system contribution}, we construct a new real-world \threesixty piece-wise planar benchmark, which focuses on evaluating horizontal and vertical planes.
Finally, our approach outperforms the two strong baselines adapted from existing state-of-the-art planar reconstruction methods on the new benchmark.
%%%%%%%%% Introduction end

%%%%%%%%% Related work start
\section{Related work}
Reconstructing 3D planes from an image involves two subtasks: segmenting plane instances and estimating plane geometric parameters.
To solve the problems, PlaneNet~\cite{LiuYCYF18} trains CNN and DCRF that reconstruct a fixed number of planes by estimating plane parameters and plane segmentation masks both in an instance-wise manner.
PlaneRecover~\cite{YangZ18} also predicts a fixed number of planes but learns directly from depth modality with plane structure-induced loss.
Recent state-of-the-art approaches relax the constraint on the number of planes by exploiting popular frameworks in instance segmentation.
PlaneRCNN~\cite{LiuKGFK19} modifies the two-stage architecture Mask R-CNN~\cite{HeGDG17} with object category classification replaced by plane geometry prediction, followed by a network to refine the segmentation masks.
PlaneAE~\cite{YuZLZG19} predicts per-pixel plane parameters and adopts associative embedding~\cite{BrabandereNG17,FathiWRWSGM17,KongF18a,NewellHD17}, which trains a network to map each pixel to embedding space and then clusters the embedded pixels to generate instances.
DualRPN~\cite{JiangLSWC20} groups planes into object and layout categories, each with its network branch, and infers plane representations for both the visible and occlusion parts.
A plane post-refinement method~\cite{QianF20} is recently proposed, which improves the results of the existing methods (\eg, PlaneRCNN and PlaneAE) by refining the inter-planar relationship.

Previous works rarely use the plane geometry prior when segmenting plane instances.
Plane parameter estimation may correlate with instance segmentation either via loss~\cite{LiuYCYF18,YangZ18,YuZLZG19} or by an additional module for refinement~\cite{LiuKGFK19,QianF20}.
In contrast, our method directly groups pixels based on plane orientations so that the plane segmentation module can detect unique planes in each group separately.

Estimating per-pixel surface normals for vertical planes from an equirectangular image is challenging for CNNs.
Specifically, vertical-plane parameters depend on \threesixty camera's yaw rotation, but the counterpart left-right circular shifting on the equirectangular image is less discerned by the translation-invariant CNNs.
Although the surface normal is a fundamental property to many \threesixty applications aside from the planar reconstruction, existing methods~\cite{YangZ16,SongZCSSF18,XuSKT17} which estimate surface normal from an equirectangular image are less aware of the \threesixty camera yaw ambiguous problem.
A workaround by~\cite{EderMG19} is to use CoordConv~\cite{LiuLMSFSY18} to make the model condition on image $u$-coordinates so that the yaw ambiguous problem in $360\degree$ is alleviated.
However, this relies on the deep net capability to learn the relationship between the $u$-coordinates and the plane orientations.
On the contrary, we propose a yaw-invariant parameterization for vertical planes, which solves the yaw-rotation ambiguous problem adequately.
%%%%%%%%% Related work end

%%%%%%%%% Dataset start
\section{PanoH\&V dataset} \label{sec:dataset}
In this section, we first introduce the large-scale panoramic public datasets used to construct our dataset (Sec.~\ref{ssec:pano_dataset_src}).
We then show the statistical analysis on these datasets to support the validity of scene approximation with H\&V-planes (Sec.~\ref{ssec:hvplane_approx}).
Finally, we outline the automatic H\&V-plane annotation algorithm from depth modality (Sec.~\ref{ssec:hvplane_extract}).

\subsection{Panorama dataset sources} \label{ssec:pano_dataset_src}
We construct our dataset from three public \threesixty RGB-D datasets, including Matterport3D~\cite{ChangDFHNSSZZ17} and Stanford2D3D~\cite{ArmeniSZS17} in real-world scenes, and Structure3D~\cite{ZhengZLTGZ19} in synthetic environments. These three datasets consist of large-scale aggregation of panoramic RGB images and depth maps. 
All panorama images in this work are represented in the equirectangular format with resolution of $512 \times 1024$.
Similar to~\cite{LiuKGFK19,LiuYCYF18} deriving plane modality to train the learning-based method, our annotations of H\&V-planar masks and parameters are derived from the ground-truth $360\degree$ depths.

We assume all panorama images are aligned with the gravity direction.
In case that g-sensor and tripod are not equipped with the \threesixty camera, and the image is not aligned, we can use the voting-based algorithm mentioned in~\cite{FernandezPLJ18,YangJLKY18,ZhangSTX14,ZouCSH18} for vanishing point (VP) detection and panoramic image alignment.
Panoramas generally provide enough context to extract the gravity direction, and we will not lose any pixel or introduce any padding after image alignment.

\subsection{H\&V-plane scene approximation} \label{ssec:hvplane_approx}
\begin{figure}
    \centering
    \includegraphics[width=0.9\linewidth]{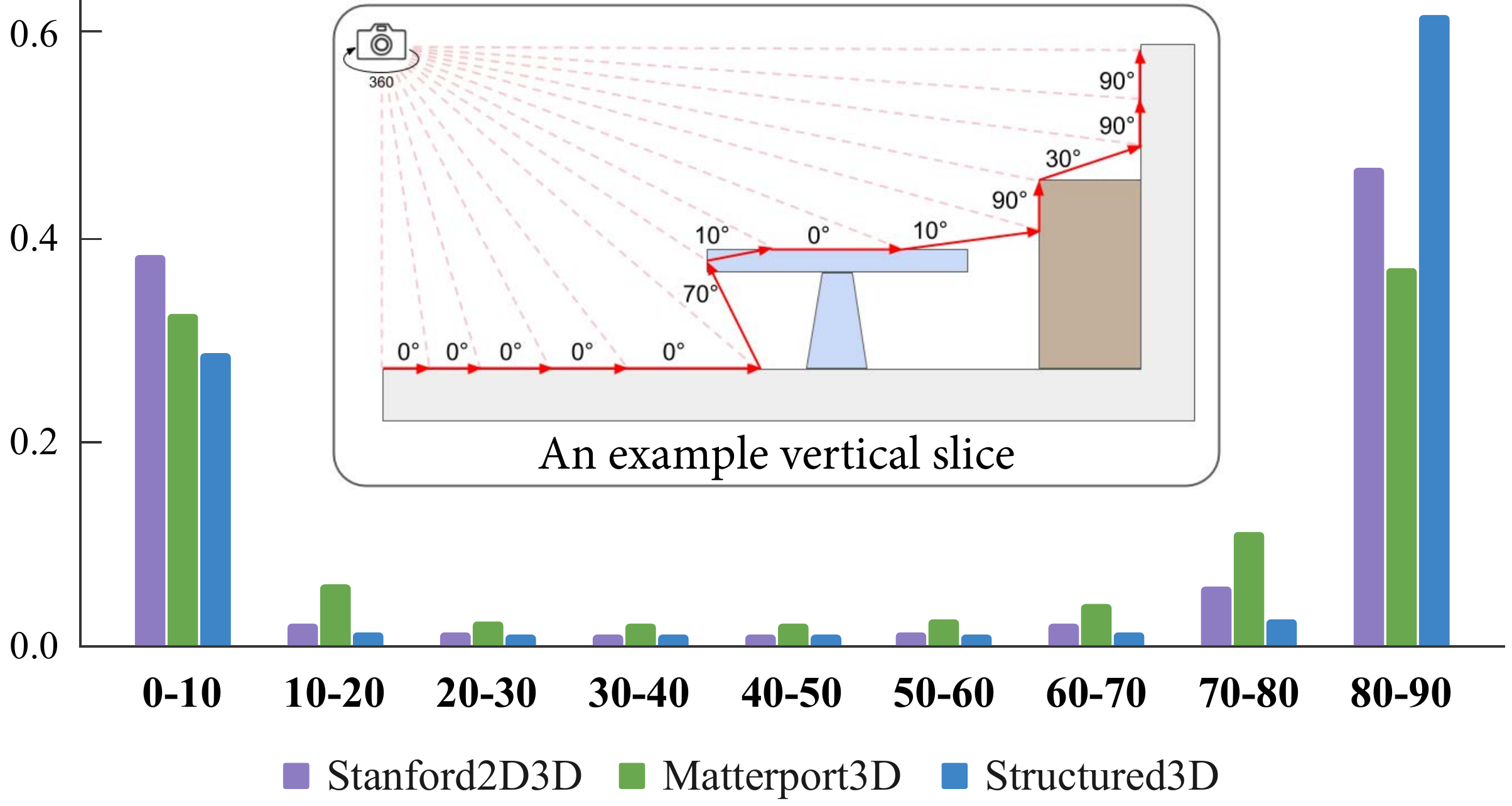}
    \caption{
    The histogram of angles between the floor plane and the line connecting two vertically adjacent pixels (both transformed into 3D points with ground truth depth).
    The center subfigure illustrates the angles of points in a vertical slice.
    The analysis shows that gravity-aligned structures dominate the indoor datasets, supporting us to approximate a scene with horizontal and vertical planes (H\&V-planes).
    }
    \label{fig:dataset_deg}
\end{figure}

\begin{figure}
  \centering
  \begin{subfigure}[t]{0.49\linewidth}
    \centering
    \includegraphics[width=0.95\linewidth]{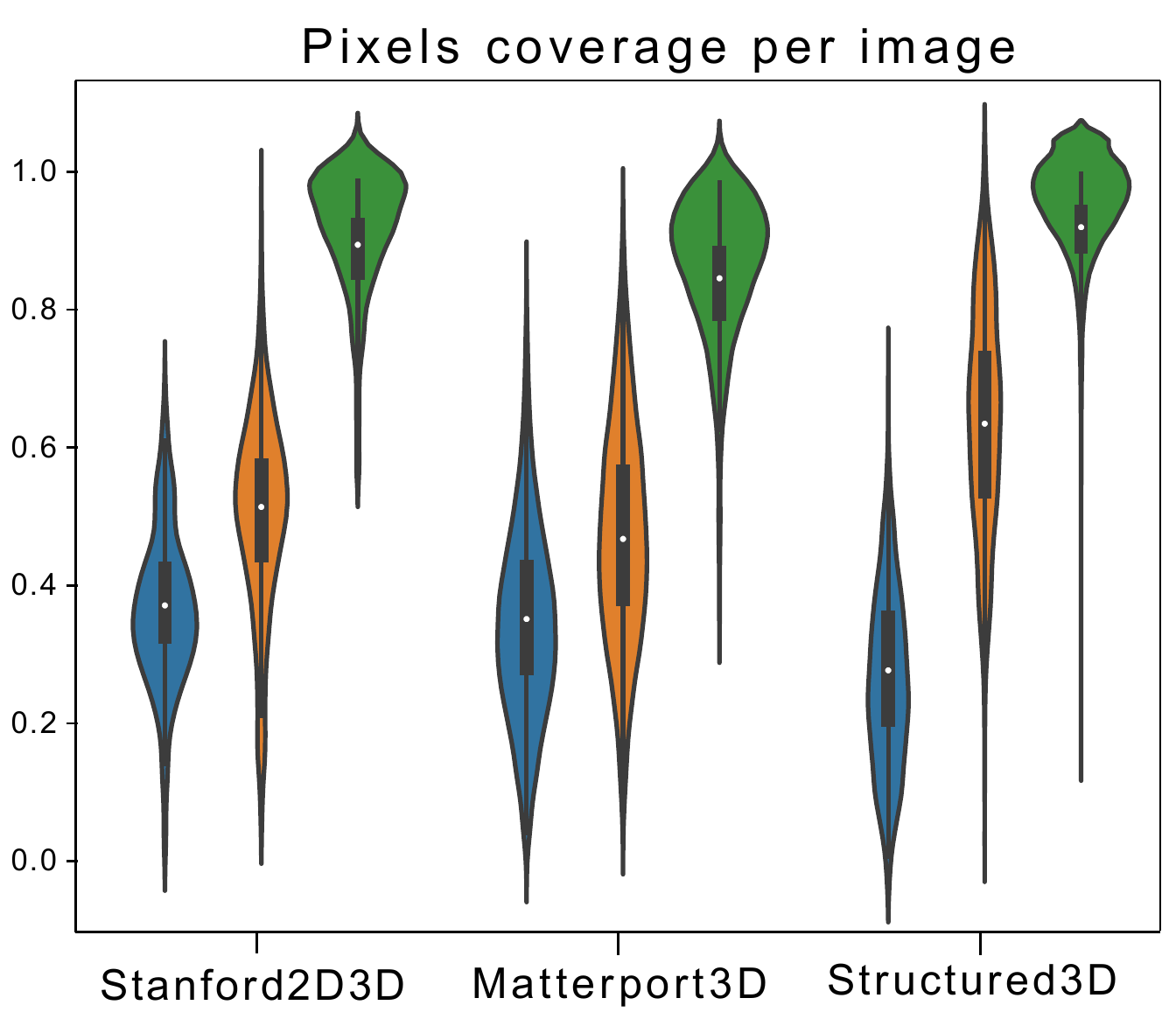}
  \end{subfigure}
  \hfill
  \begin{subfigure}[t]{0.49\linewidth}
    \centering
    \includegraphics[width=0.95\linewidth]{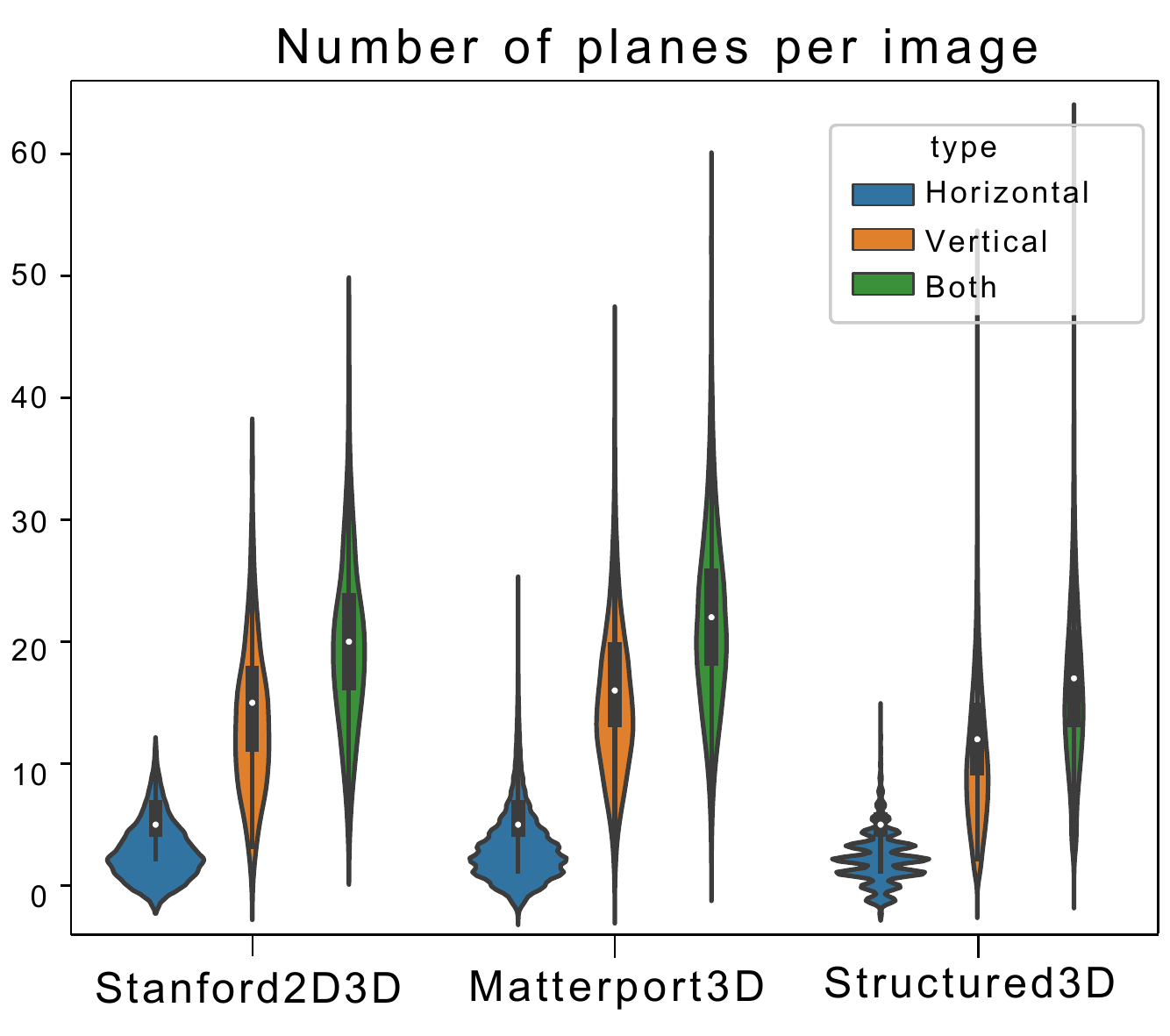}
  \end{subfigure}
  \caption{
    The violin plot illustrates the statistics of the constructed H\&V-plane benchmark.
    We show all the combinations between the three panoramic datasets ($x$-axis) and the type of planes (blue bars for H-planes, orange bars for V-planes, and green bars for both).
    The left canvas shows the per-image pixel coverage of H\&V-planes. The $y$-axis is the ratio of covered pixels.
    The right canvas shows the number of H\&V-planes per image.
  }
  \label{fig:dataset_statistic}
\end{figure}

We aim to approximate the indoor scene by horizontal planes (\emph{H-planes}) and vertical planes (\emph{V-planes}) whose surface normals are parallel and perpendicular to the gravity direction respectively.
H\&V-plane is a restrictive version of the general
piecewise plane and can fit the Manhattan world~\cite{CoughlanY00} (MW) and the Atlanta world~\cite{SchindlerD04} (AW), but, unlike MW and AW, the H\&V-planes do not assume any relationship between V-planes.
Although the use of H\&V-planes for scene approximation complies with our intuition, we further perform two quantitative analyses on three large-scale indoor panorama datasets, which include various indoor scenes (\eg, classroom, office, living room, kitchen) to justify the validity and applicability of H\&V-plane assumption.
\begin{enumerate}[label=\arabic*),leftmargin=*]
  \item For each pair of vertically adjacent pixels in aligned \threesixty images, we calculate the angle between the floor plane and the line passing through the two projected 3D points.
  The histogram of angles computed from all images of the three indoor panoramic datasets is shown in Fig.~\ref{fig:dataset_deg}.
  Imagine that we take a vertical slice of the 3D point cloud (corresponding to an image column) and move from the bottom point to its upper adjacent point and repeat the process until we reach the top point---Most of the moving directions will be either horizontal (0$\degree$ to 10$\degree$) or vertical (80$\degree$ to 90$\degree$).
  Such distribution suggests the gravity aligned nature of indoor scene structures.
  \item We approximate the scenes projected from the depth modality with our H-planes and V-planes extraction algorithms (described in Sec.~\ref{ssec:hvplane_extract}).
  The statistics of the derived H\&V-plane annotation on the three indoor panoramic datasets are shown in Fig.~\ref{fig:dataset_statistic}.
  We can see that, in general, more than 80\% of the pixels in an image can be covered (less than 5cm discrepancy) by roughly 20 H\&V-planes. This result suggests that our H\&V-plane approximation are suitable for modeling the gist of an indoor scene. 
\end{enumerate}


\subsection{Ground-truth H\&V-plane annotation} \label{ssec:hvplane_extract}
Following the gravity aligned and H\&V-plane assumptions, we present the first \threesixty planar dataset with training, validation, and test sets.
The annotations include H\&V-plane masks and plane parameters, and all images and annotations are in the same resolution of $512 \times 1024$.
We use the same analysis as in Fig.~\ref{fig:dataset_deg} to classify each pixel into H-pixel, V-pixel, or other.
RANSAC is then performed on H-pixels and V-pixels to extract instance mask and plane geometry for H-planes and V-planes.
The statistical information of the constructed dataset is depicted in Fig.~\ref{fig:dataset_statistic}.
Please refer to supplementary material for the full description of the H\&V-plane extraction algorithm.
%%%%%%%%% Dataset end

\begin{figure*}
  \centering
  \includegraphics[width=0.8\linewidth]{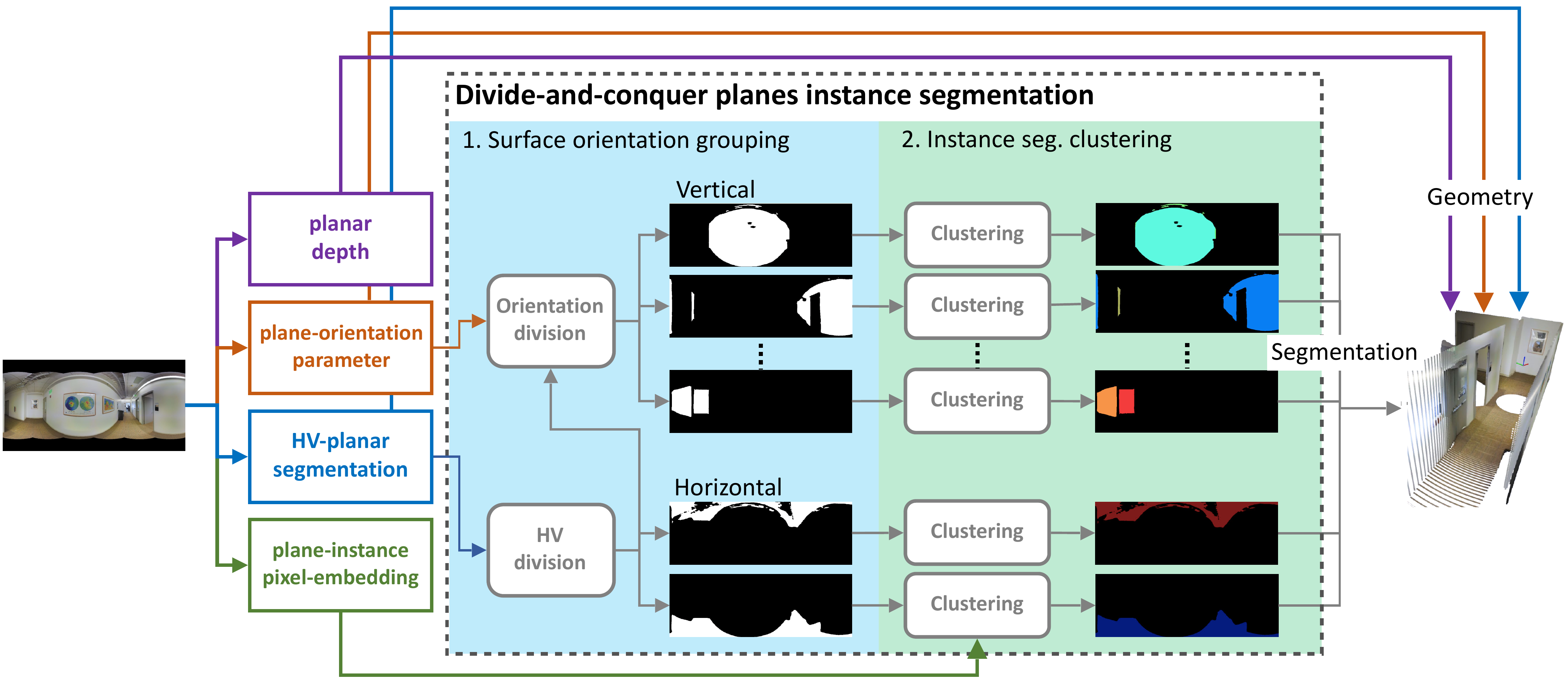}
  \caption{
    An overview of the proposed method.
    We employ an encoder-decoder network to extract pixel-level latent features from the input RGB panorama, followed by different $\operatorname{Conv1x1}$ layers to produce each modality.
    We detail the HV-planar segmentation and geometric characteristics estimation in Sec.~\ref{ssec:hvplanar} and Sec.~\ref{ssec:geometry} respectively.
    In Sec.~\ref{ssec:instseg}, a divide-and-conquer strategy is proposed exploiting all the pixel-level predictions for plane instance segmentation.
    We first \emph{divide} pixels according to the estimated surface orientation.
    In the \emph{conquer} stage, we employ associative embedding~\cite{YuZLZG19} to distinguish individual plane instances in an orientation group, which only involves pixels of a similar plane orientation and thus easier to solve.
  }
  \label{fig:model}
  \vspace{-1em}
\end{figure*}

%%%%%%%%% Apporach start
\section{Approach}

Our task is to reconstruct H\&V-planes from a single \threesixty RGB image.
An overview of our approach is depicted in Fig.~\ref{fig:model}.
Our first step is to partition an image into H-planar, V-planar, and non-planar regions (Sec.~\ref{ssec:hvplanar}).
Our model predicts planar depth and V-plane orientation both in pixel-level, which will be used to derive the plane parameters (Sec.~\ref{ssec:geometry}).
In Sec.~\ref{ssec:instseg}, a divide-and-conquer strategy for plane instance segmentation is proposed to make use of the learned pixels' H\&V-planar and geometric information.
Finally, implementation details are provided in Sec.~\ref{ssec:arch_train_detail}.

\subsection{H\&V-planar segmentation} \label{ssec:hvplanar}
We train two binary classifiers to be activated on H-planar pixels and V-planar pixels respectively.
A pixel is recognized as ``non-planar'' when the corresponding probabilities from both classifiers are below a certain threshold (we simply set to $0.5$); otherwise, it will be classified as H-pixels or V-pixels according to which classifier gives a higher probability.
An alternative is to use 3-way softmax classification, but we do not observe performance differences and thus stick to two binary classifiers with fewer output channels.
We use binary cross entropy loss ($\operatorname{BCE}$) as H\&V-planar training objective:
\begin{equation}
L_{\mathrm{HVseg}} = \frac{1}{N} \sum_{i} \big(\operatorname{BCE}\,(m_i^{h}, m_i^{h*}) + \operatorname{BCE}\,(m_i^{v}, m_i^{v*})\big) ,
\end{equation}
where the subscript $i$ is the pixel index, $N$ is the total number of pixels, $\{m^{h}, m^{v}\}$ are the predicted H\&V-planar probability maps, and $\{m^{h*}, m^{v*}\}$ are the ground truth masks.
Based on the segmented H\&V-pixels, we exploit their prior to give different treatments in the succeeding modules.

\subsection{Planar geometry estimation} \label{ssec:geometry}
Our model estimates two geometric characteristics at pixel-level---H\&V-planar depth $d$ and V-planar surface normal $\theta$.
The H\&V-planar depth $d$ is used to recover both H-planes and V-planes, while the V-plane orientation $\theta$ only relates to V-planes.
In below, the plane parameter is denoted as $\vec{n} = [x\; y \;z]$, which is the unit surface normal scaled by the plane offset; image coordinate is denoted as $u \in [-\pi, \pi]$ and $v \in [-\frac{\pi}{2}, \frac{\pi}{2}]$; we use subscript $i$ to denote the index of a pixel.
In testing phase, the plane parameter of a detected plane is simply determined by the dimension-wise median of $\vec{n}_i$ in the segmented instance mask.

\subsubsection{H-planar geometry} \label{sssec:hplanar_geo}
The unit surface normal of an H-plane is either $[0 \; 0 \; 1]$ or $[0 \,\; 0 \; -1]$ (corresponding to horizontal planes above or below the camera respectively) and can be determined accordingly as the pixel is located at the upper half or the bottom half of an equirectangular image.
Thus, the only one degree-of-freedom of an H-plane is the plane offset.
Given the estimated planar depth $d$, we derive the H-plane offset at the pixel-level by
\begin{equation}
z_i = d_i \cdot \sin v_i \,.
\end{equation}
Thus, $\vec{n}_i = z_i \cdot [0 \,\; 0 \,\; 1]$ is the plane parameter of the H-planar pixel, where the sign is determined by $\sin\,(v_i)$.
The training objective for H-planar geometry is
\begin{equation}
L_{\mathrm{Hgeo}} = \frac{1}{|I_H|} \sum_{i\in I_H} |z_i - z_i^*| \,,
\end{equation}
where $I_H$ is the set of all H-pixels indices and $z_i^*$ is the ground truth H-plane offset of the pixel.

\subsubsection{V-planar geometry} \label{sssec:vplanar_geo}
For a pixel belonging to a V-plane, we derive the V-plane surface normal from the estimated $\theta_i$ as $[\cos\theta_i \; \sin\theta_i \; 0]$.
The V-plane offset depends on the V-plane surface normal and the estimated planar depth $d_i$, which is
\begin{equation}
o_i = d_i \cdot \cos v_i \cdot [\cos\theta_i \; \sin\theta_i] \cdot [\cos u_i \; \sin u_i]^T .
\end{equation}
Thus, $\vec{n}_i = o_i \cdot [\cos\theta_i \; \sin\theta_i \; 0]$ is the plane parameter of the V-planar pixel.
Note that only $\theta$ and $d$ are model predictions, while $o$ and $\vec{n}$ are derived from $\theta$ and $d$.
The V-planar loss is
\begin{equation}
L_{\mathrm{Vgeo}} = \frac{1}{|I_V|} \sum_{i \in I_V} \| \vec{n}_i - \vec{n}_i^*\|_1 + \operatorname{CosineLoss}\,(\theta_i, \theta_i^*) ,
\end{equation}
where $\vec{n}_i^*$ and $\theta_i^*$ are ground-truth V-planar geometries and 
\begin{equation}
\operatorname{CosineLoss}\,(\theta_i, \theta_i^*) = 1 - (\cos\theta_i \cos\theta_i^* + \sin\theta_i \sin\theta_i^*) .
\end{equation}

\paragraph{Yaw-invariant parameterization for V-plane orientations.}
The accuracy of the estimated V-plane orientation $\theta$ is critical in our method.
It affects the planar geometry quality and also involves in the proposed divide-and-conquer process of plane instance segmentation (Sec.~\ref{ssec:instseg}).
However, V-plane orientation estimation in a \threesixty image is a challenging task for CNNs.
The reason is that the V-planar surface orientation co-varies with the \threesixty camera yaw rotation, but the translation-invariant CNNs are less aware of the corresponding left-right circular shifting on the equirectangular image.
Employing CoordConv~\cite{LiuLMSFSY18} to condition the CNNs on image coordinate is a workaround for this issue.

To address the yaw ambiguity more effectively, we propose re-parameterizing $\theta_i$ into residual form with respect to the pixel yaw viewing angle $u_i$ such that it is invariant to the \threesixty camera yaw rotation.
More specifically, instead of appending $u_i$ to input like CoordConv~\cite{LiuLMSFSY18}, we subtract it from the target orientation:
\begin{equation}
{\theta'}_i^* = \theta_i^* - u_i \,,
\end{equation}
which is the re-parameterized yaw-invariant V-plane orientation.
The proposed representation enables the model to infer the V-plane orientation without the knowledge about the \threesixty camera yaw rotation.

\subsection{Divide-and-conquer for plane instance segmentation} \label{ssec:instseg}

Motivated by the strong prior in indoor scenes where most plane instances share a small number of distinct orientations (\eg, Manhattan world~\cite{CoughlanY00} and Atlanta world~\cite{SchindlerD04}), we propose to integrate plane orientation information into the main process of plane instance segmentation.
We use a divide-and-conquer strategy.
In the \emph{divide} stage, we divide pixels by surface orientation grouping, which forms multiple simpler subproblems.
In the \emph{conquer} stage, we apply pixel embedding clustering~\cite{YuZLZG19} to identify unique plane instances in each orientation group.

We find a similar idea in the recent DualRPN~\cite{JiangLSWC20}, where the planes are separated into two predefined groups---object and layout---each processed by its branch.
In contrast to DualRPN, we do not need semantic annotation for our \emph{divide} stage, and our orientation groups are automatically determined, where there are more than six groups in most of the cases (\eg, the Manhattan world captured by panorama).

\subsubsection{\emph{Divide}: surface orientation grouping}
\begin{figure}
  \centering
  \begin{subfigure}[b]{0.65\linewidth}
    \centering
    \includegraphics[width=\linewidth]{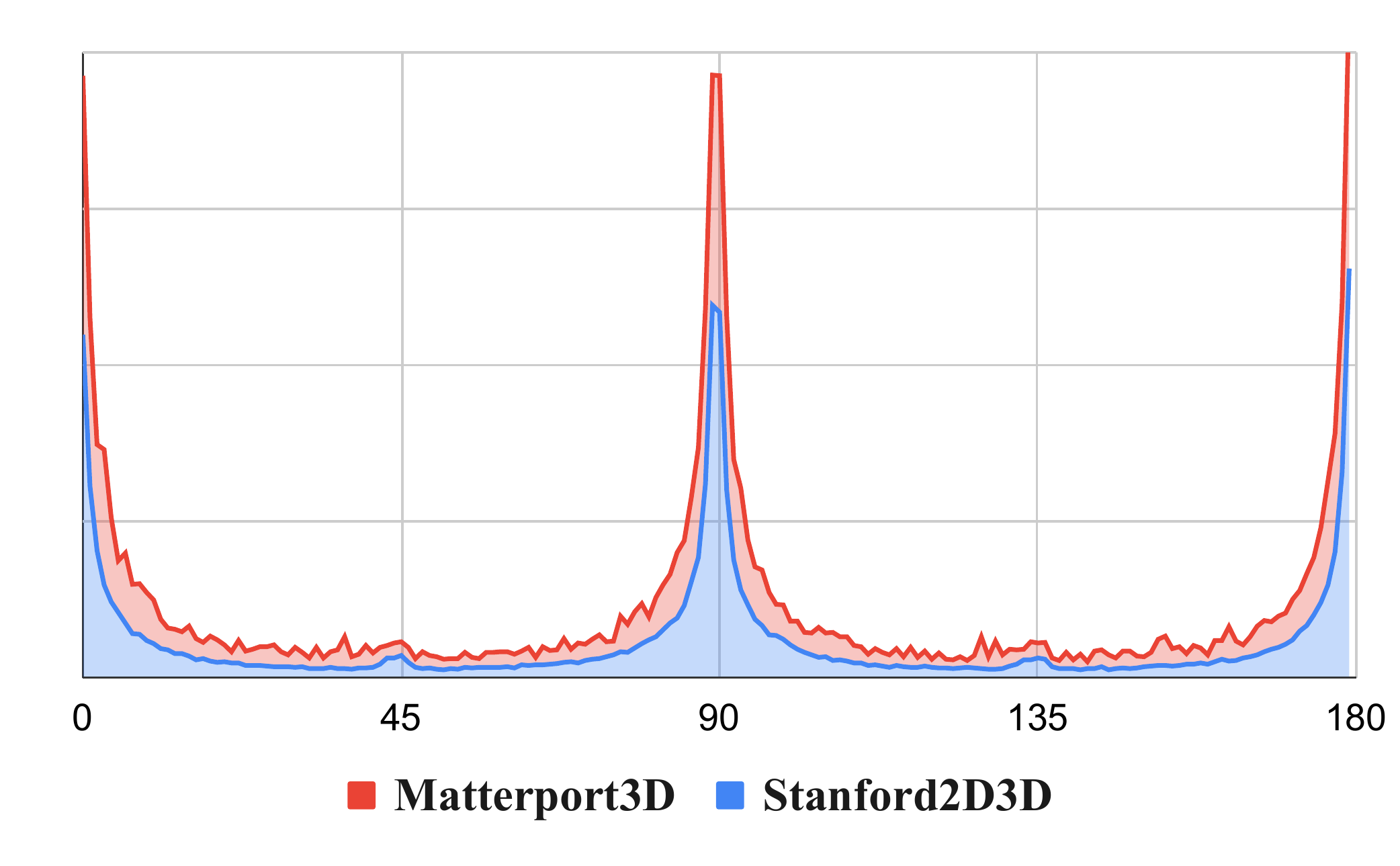}
    \caption{}
  \end{subfigure}
  \hfill
  \begin{subfigure}[b]{0.32\linewidth}
    \centering
    \includegraphics[width=\linewidth]{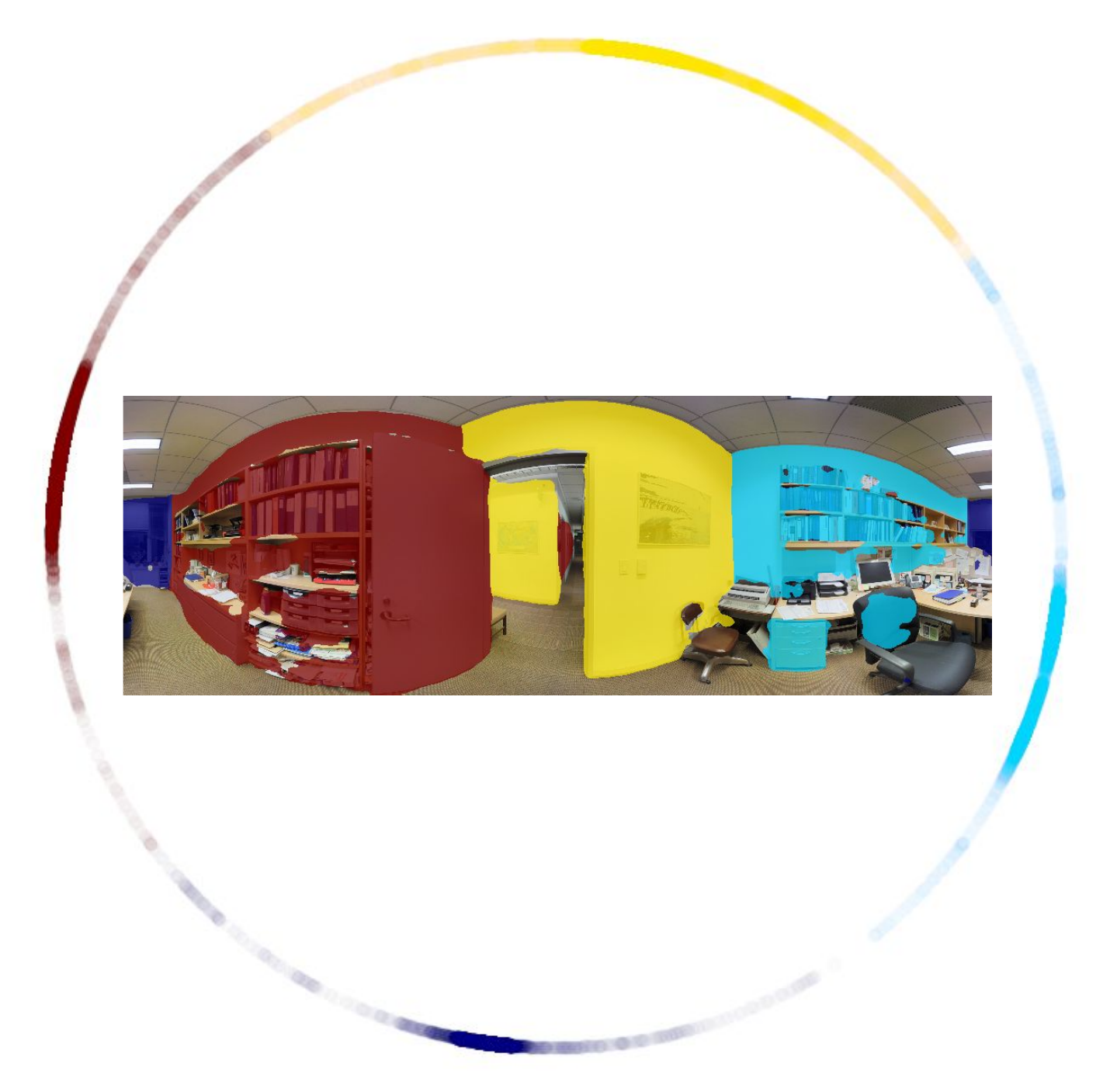}
    \caption{}
    \label{fig:preseg_vis}
  \end{subfigure}
  \caption{
    (a) Statistics of angles between all pairs of planes in an image. The results are averaged across the entire Stanford2D3d and subsampled Matterport3D.
    (b) An example of surface orientation grouping result. Different colors denote different groups. The outer circle represents the distribution of the estimated V-plane orientations.
  }
  \label{fig:yaw_deg}
  \vspace{-1em}
\end{figure}

The analysis on our dataset (see Fig.~\ref{fig:yaw_deg}) shows that, in an indoor panorama, most plane instances share similar, opposite, or perpendicular plane orientations with other planes.
Hence, the per-pixel V-planar surface orientations are distributed primarily around a small number of angles.
Utilizing such regularity, we divide pixels of an image into groups of similar plane normals to early separate plane instances of dissimilar orientations.
We first use the predicted H\&V-planar mask (Sec.~\ref{ssec:hvplanar}) for dividing H\&V-pixels (and ignoring ``non-planar'' pixels).
For H-pixels, the two surface orientation groups (\ie $\vec{n} = [0 \;\, 0 \; \pm 1]$) can be easily determined by pixels' $v$-coordinates (Sec.~\ref{sssec:hplanar_geo}).
For V-pixels, we use a voting process to detect the prominent V-plane orientation peaks.
Specifically, we quantize the estimated $\theta$ of V-pixels into circular bins followed by a peak-finding algorithm (see supplementary material for more detail).
V-pixels are assigned to their nearest peak to form surface orientation groups (see Fig.~\ref{fig:preseg_vis} for an example in practice).

\subsubsection{\emph{Conquer}: pixel embedding clustering}
To further cluster pixels in each surface orientation group into plane instance masks, we employ pixel embedding, which is originally dealing with instance segmentation~\cite{BrabandereNG17,FathiWRWSGM17,KongF18a} and also found to be effective for plane instance segmentation~\cite{YuZLZG19}.
We follow the setting in \cite{YuZLZG19} for the pixel embedding---embedding dimension set to $2$; trained with losses $L_{\mathrm{pull}}$ to pull the pixels of a plane instance toward their centroid and $L_{\mathrm{push}}$ to push the centroids of different planes away from each other.
In this work, we use different pixel embedding spaces for H-planes and V-planes, so $L_{\mathrm{pull}} = L_{\mathrm{Hpull}} + L_{\mathrm{Vpull}}$ and $L_{\mathrm{push}} = L_{\mathrm{Hpush}} + L_{\mathrm{Vpush}}$.
In the testing phase, the mean shift clustering is applied to each orientation group separately.

\subsection{Implementation details} \label{ssec:arch_train_detail}
We extract pixel-level latent features by employing the standard encoder-decoder architecture.
In below, we use $\operatorname{ConvBlock}$ to denote a sequence of $\operatorname{Conv3x3}$, $\operatorname{BN}$, and $\operatorname{ReLU}$.
We use ResNet-101~\cite{HeZRS16} as our backbone.
The channels of each ResNet's output features are first reduced to $128$ by two $\operatorname{ConvBlocks}$.
In the decoder, the features from a lower resolution are upsampled and concatenated with higher resolution features, followed by a $\operatorname{ConvBlock}$ to fuse the two sources of features.
In the upsampling process, the channels remain $128$ except the finest scale (spatial stride $2$), whose channels are set to $64$.
Finally, each modality is produced by a $\operatorname{Conv1x1}$ layer from the finest scale features.
All the predictions are upsampled by $2$ to the input resolution.
See supplementary material for the architecture diagram.
The overall training objective is
\begin{equation}
L = 0.1\,L_{\mathrm{HVseg}} + L_{\mathrm{Hgeo}} + L_{\mathrm{Vgeo}} + L_{\mathrm{pull}} + L_{\mathrm{push}} ~ .
\end{equation}
We give the H\&V-planar segmentation loss $L_{\mathrm{HVseg}}$ a smaller weight as we find it converges quickly after just a few epochs.
%%%%%%%%% Approach end

\begin{table*}
    \centering
    \begin{tabular}{l|ccc|cc|ccc}
        \hline
        \multirow{2}{*}{Method} &
        \multicolumn{3}{c|}{Segmentation Quality} &
        \multirow{2}{*}{Per-pixel Recall $\uparrow$} &
        \multirow{2}{*}{Per-plane Recall $\uparrow$} & \multicolumn{3}{c}{Depth accuracy}\\
         & ARI$\uparrow$ & VI$\downarrow$ & SC$\uparrow$ &  &  & Rel.$\downarrow$ & $\log_{10} \downarrow$ & RMSE$\downarrow$ \\
        \hline
        \hline
        \multicolumn{6}{l}{\textbf{Matterport3D~\cite{ChangDFHNSSZZ17} test set}} \\
        \hline
        PlaneRCNN~\cite{LiuKGFK19} &
        0.574 & 2.022 & 0.632 & 0.473 & 0.336 & {\bf 0.125} & {\bf 0.052} & 0.329 \\
        \hline
        PlaneAE~\cite{YuZLZG19}  & 
        0.673 & 1.944 & 0.640 & 0.498 & 0.381 & 0.187 & 0.080 & 0.528 \\
        \hline
        Ours &
        \textbf{0.686} & \textbf{1.894} & \textbf{0.660} & \textbf{0.544} & \textbf{0.410} & 0.132 & 0.053 & {\bf 0.326} \\
        \hline
        \hline
        \multicolumn{6}{l}{\textbf{Stanford2D3D~\cite{ArmeniSZS17} validation set}} \\
        \hline
        PlaneRCNN~\cite{LiuKGFK19}  & 
        0.682 & 1.677 & 0.703 & 0.452 & 0.297 & 0.119 & 0.055 & 0.386\\
        \hline
        PlaneAE~\cite{YuZLZG19}  & 
        0.765 & 1.536 & 0.733 & 0.520 & 0.341 & 0.140 & 0.071 & 0.671\\
        \hline
        Ours &
        \textbf{0.768} & \textbf{1.514} & \textbf{0.742} & \textbf{0.627} & \textbf{0.430} & \textbf{0.093} & \textbf{0.041} & \textbf{0.327} \\
        \hline
        \hline
        \multicolumn{6}{l}{\textbf{Structured3D~\cite{ZhengZLTGZ19} test set}} \\
        \hline
        PlaneRCNN~\cite{LiuKGFK19}  & 
        0.726 & 1.393 & 0.743 & 0.654 & 0.522 & 0.071 & 0.029 & 0.137 \\
        \hline
        PlaneAE~\cite{YuZLZG19}  & 
        0.821 & 1.175 & 0.785 & 0.728 & 0.591 & 0.122 & 0.054 & 0.464 \\
        \hline
        Ours &
        \textbf{0.824} & \textbf{1.150} & \textbf{0.794} & \textbf{0.794} & \textbf{0.657} & \textbf{0.057} & \textbf{0.025} & \textbf{0.126} \\
        \hline
    \end{tabular}
    \caption{
        The new benchmark.
        Our method outperforms the two adapted strong baselines.
    }
    \label{table:quan_test}
    \vspace{-2em}
\end{table*}

%%%%%%%%% Experiment start
\section{Experiments} \label{sec:exp}

\subsection{Baselines construction}
We manage to adapt two current state-of-the-art planar reconstruction approaches--- PlaneAE~\cite{YuZLZG19}\footnote{\url{https://github.com/svip-lab/PlanarReconstruction}} and PlaneRCNN~\cite{LiuKGFK19}\footnote{\url{https://github.com/NVlabs/planercnn}}---with their official implementation to our newly constructed benchmark.
For simplicity and consistency to ours, the input panorama format is equirectangular with resolution $512 \times 1024$, and we make necessary changes to the baseline methods accordingly.
Using cubemap as the input format is an alternative, but more detailed designs should be considered for the separated faces (\eg, feature delivering strategy, crossing-multi-faces bounding boxes merging or masks merging strategy), which is out of scope for this work.
Observing the success of processing equirectangular with planar CNNs in estimating depth~\cite{JinXZZTXYG20} and layout~\cite{ZouCSH18,SunHSC19}, we believe fine-tuning the ImageNet~\cite{DengDSLL009} pre-trained planar CNNs on equirectangular is suitable for our application.

\paragraph{Common adaptation.}
We employ the left-right circular padding~\cite{WangHLHZS18} for all convolution layers.
The $u$-coordinate is concatenated as one of the input channels to alleviate the ambiguity of yaw rotation.
All methods use ResNet-101~\cite{HeZRS16} as the backbone.
The unit surface normals have two degree-of-freedom $\theta_u, \theta_v$, where $\theta_v$ in our case is produced by a binary classifier as they are either $0$ or $\pm1$ (the sign can be determined by pixel vertical location).

\paragraph{PlaneRCNN~\cite{LiuKGFK19}.}
We modify RoIAlign and NMS to handle the left-right circular coordinate system.
We have normal clusters for vertical and horizontal planes.

\paragraph{PlaneAE~\cite{YuZLZG19}.}
The Efficient Mean Shift is scaled according to the image resolution---number of iterations, anchors and sampled points are set to $18$, $36$, and $11$k respectively.

\subsection{Data split}
We follow the official to split the scenes into training, validation, and test sets, but remove data that have too many missing depth values while extracting the ground-truth H\&V-planes (Sec.~\ref{sec:dataset}).
Finally, Stanford2D3D~\cite{ArmeniSZS17} contains 1{,}040 images for training and 372 images for validation; Matterport3D~\cite{ChangDFHNSSZZ17} contains 7{,}275, 1{,}189, and 1{,}005 for training/validation/test; Structured3D~\cite{ZhengZLTGZ19} contains 18{,}332 for training, 1{,}771 for validation, and 1{,}691 for test.

\subsection{Training protocol}
We use Adam optimizer~\cite{KingmaB14} with learning rate 1e-4 and $2$ panoramas in a mini-batch.
The number of training epochs for Stanford2D3D, Matterport3D, and Structured3D are $100$, $20$, and $10$ respectively.
All methods, including ours, use the same training protocol.

\subsection{Evaluation metrics} \label{ssec:metric}
Following previous works~\cite{LiuKGFK19,YangZ18,YuZLZG19}, we evaluate the performance of plane instance segmentation with some standard clustering metrics~\cite{ArbelaezMFM11}: Adjusted Rand Index (ARI$\uparrow$), Variation of Information (VI$\downarrow$), and Segmentation Covering (SC$\uparrow$) which only consider the unique plane instance segmentation results on the 2D image.
To evaluate 3D reconstruction quality, plane and pixel recalls are used under different planar depth discrepancy thresholds and mask IoU over $0.5$.
We report the results by averaging the plane recall and pixel recall under depth threshold of 5cm, 10cm, 20cm, 30cm, and 60cm.
Pixel-level depth accuracy on ground truth planar region are also reported.

\subsection{Results} \label{ssec:results}
\paragraph{Quantitative evaluation.}
In Table~\ref{table:quan_test}, we report comparisons between our approach and two competitive baselines on three presented \threesixty datasets.
Our method achieves state-of-the-art performance on all planar metrics.
For the metrics that only relate to segmentation quality, our method shows more improvement on Matterport3D, which contains more complex scenes captured in luxury houses comparing to the performance gain on Stanford2D3D and the synthetic Structured3D dataset.
For metrics considering 3D reconstruction quality, our approach outperforms both baselines by a large margin.
PlaneRCNN is trained against with ground truth depth while PlaneAE and our method only learn from plane parameters.
On depth-based evaluation, our method still shows competitive results comparing to PlaneRCNN.
% The results in Table~\ref{table:quan_test} clearly demonstrate the effectiveness of the proposed method.

\begin{figure*}
  \centering
  \begin{subfigure}[t]{0.497\linewidth}
    \centering
    \includegraphics[width=.83\linewidth]{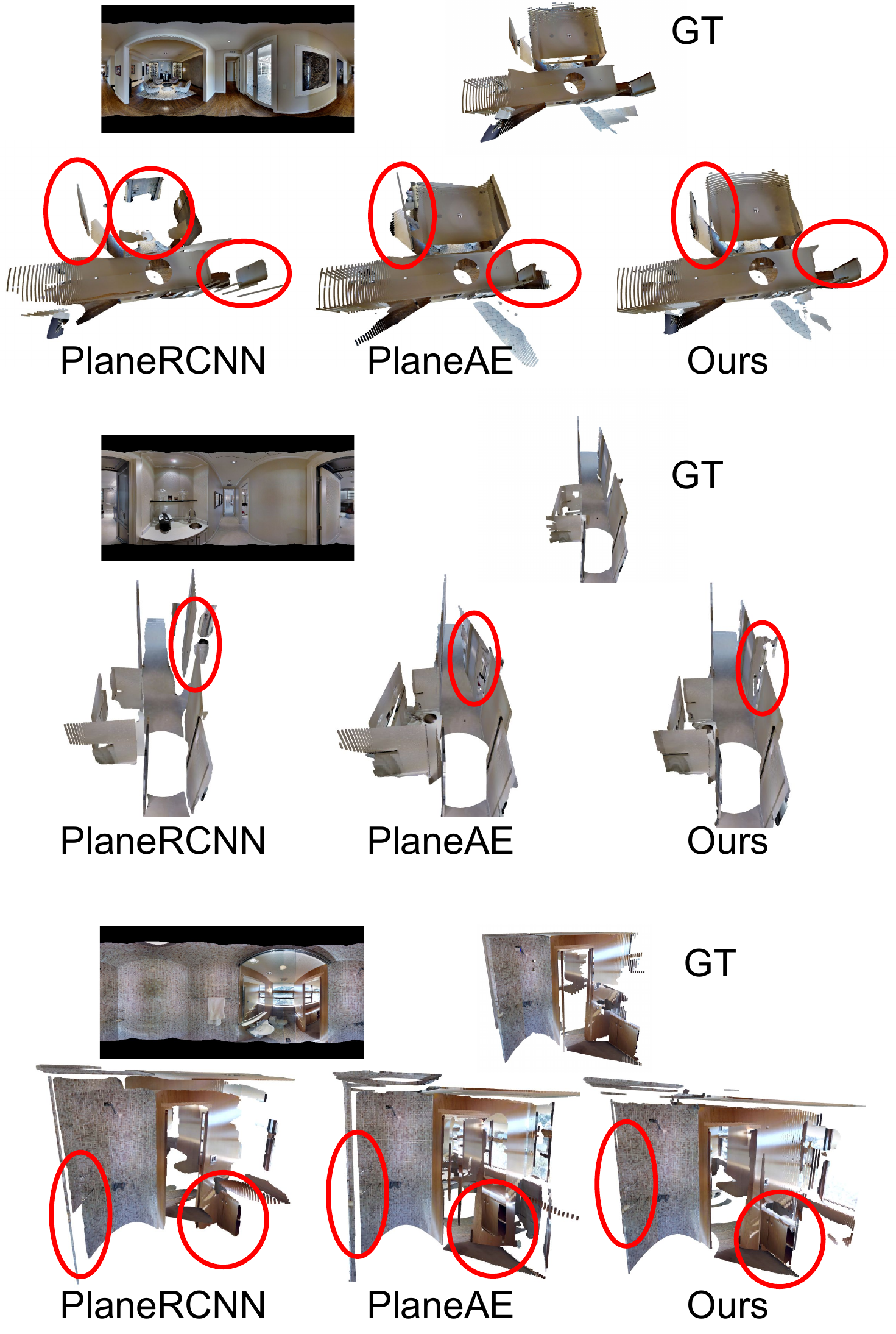}
  \end{subfigure}
  \hfill
  \begin{subfigure}[t]{0.497\linewidth}
    \centering
    \includegraphics[width=.83\linewidth]{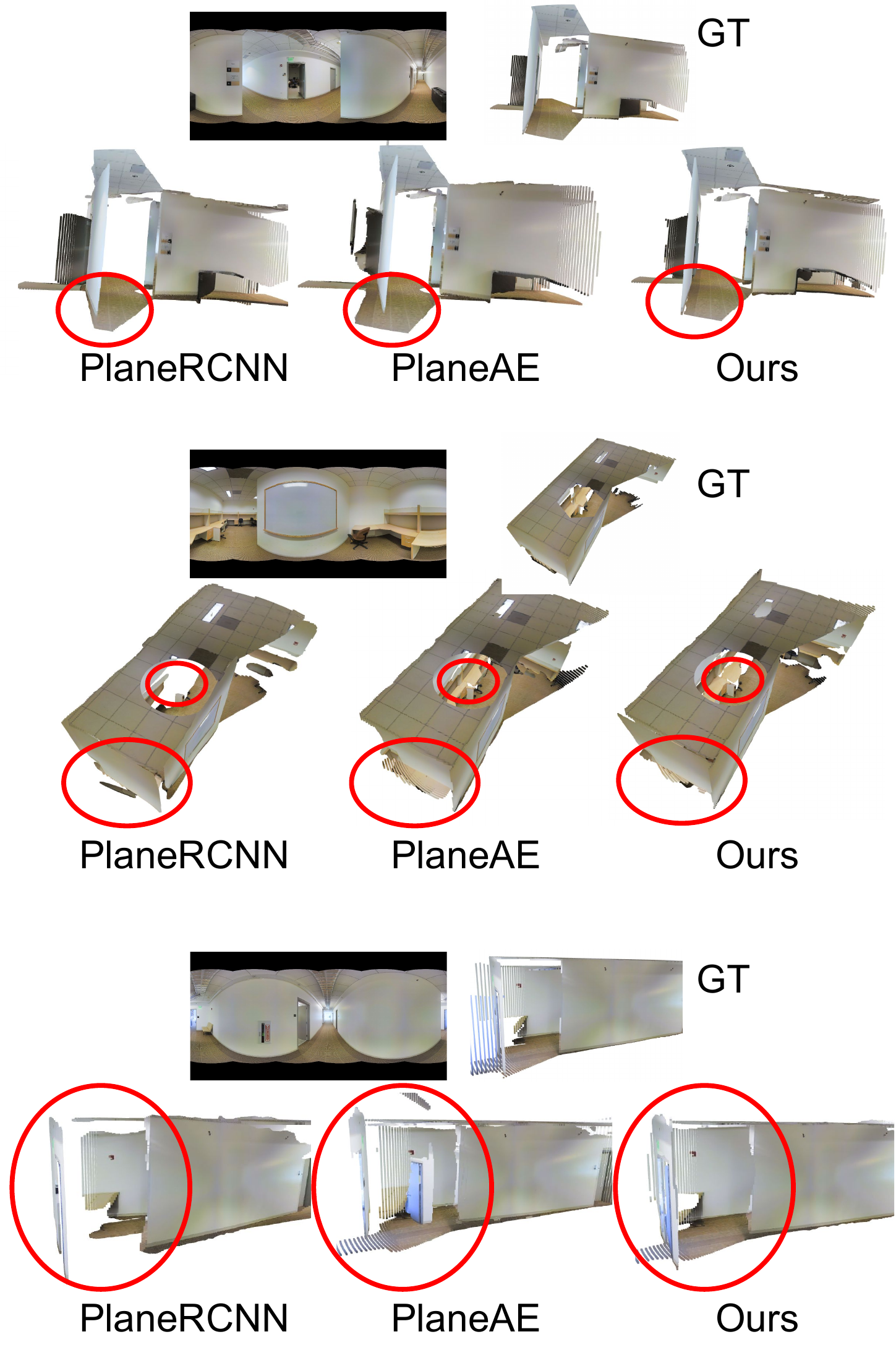}
  \end{subfigure}
  \caption{
    We show some qualitative comparisons of the 3D reconstruction results among the three methods.
    We highlight the differences in red circles.
    See supplement for more comparisons on 2D and 3D mesh visualization in bird-eye view.
  }
  \label{fig:finalqual}
  \vspace{-1em}
\end{figure*}

\paragraph{Qualitative results.}
In Fig.~\ref{fig:finalqual}, we show the reconstructed visualization by PlaneRCNN~\cite{LiuKGFK19}, PlaneAE~\cite{YuZLZG19}, and ours.
Owing to our yaw-invariant representation, our geometric quality is generally better than the others'.
In addition, our reconstructed planes are aligned better with each other.

\paragraph{Processing time.}
The average processing time over 50 scenes (with GeForce RTX 2080 Ti) for PlaneRCNN, PlaneAE and ours are 137ms, 144ms and 364ms respectively.
The proposed D\&C takes 224ms, which can be improved by applying the mean shift to each division in parallel instead of current sequential implementation.

\begin{table}
    \centering
    \begin{tabular}{c@{\hskip 0pt}c|c@{\hskip 6pt}c}
        \hline
        Method & \textbf{yaw-invariant} & \makecell{Per-pixel\\recall $\uparrow$} & \makecell{Per-plane\\recall $\uparrow$} \\
        \hline
        \hline
        \multicolumn{4}{l}{\textbf{Matterport3D~\cite{ChangDFHNSSZZ17} validation set}}\\
        \hline
        \multirow{2}{*}{PlaneRCNN~\cite{LiuKGFK19}} & & 0.471 & 0.313 \\
        \cline{3-4}
         & \checkmark & \textbf{0.495} & \textbf{0.326} \\
        \hline
        \multirow{2}{*}{PlaneAE~\cite{YuZLZG19}} &           & 0.484 & 0.362 \\
        \cline{3-4}
         & \checkmark & \textbf{0.495} & \textbf{0.369} \\
        \hline
        \hline
        \multicolumn{4}{l}{\textbf{Stanford2D3D~\cite{ArmeniSZS17} validation set}}\\
        \hline
        \multirow{2}{*}{PlaneRCNN~\cite{LiuKGFK19}} & & 0.452 & 0.297 \\
        \cline{3-4}
         & \checkmark & \textbf{0.481} & \textbf{0.313}\\
        \hline
        \multirow{2}{*}{PlaneAE~\cite{YuZLZG19}} &  & 0.520 & 0.341 \\
        \cline{3-4}
         & \checkmark & \textbf{0.555} & \textbf{0.368} \\
        \hline
    \end{tabular}
    \caption{
        Applying the yaw-invariant parameterization to the two baselines shows better planar reconstruction results.
    }
    \label{table:baseline_abla}
    \vspace{-2em}
\end{table}

\subsection{Ablation study} \label{ssec:ablation}

In this section, several ablation studies are shown to further exemplify the effectiveness of the proposed method.

\paragraph{Can the proposed \threesixty yaw-invariant plane parameterization benefit other baselines?}
One challenge of \threesixty plane orientation estimation is that the non-horizontal normals are related to camera yaw rotation, but the CNNs are less aware of the counterpart 2D left-right circular shifting on the equirectangular image.
In addition to CoordConv~\cite{LiuLMSFSY18} stategy, we propose a \threesixty camera yaw-invariant representation for V-plane normals, which ensures the ground truth normal is indepent of the ambiguous camera yaw rotation (see Sec.~\ref{ssec:geometry} for detail).
The proposed representation is applied to other baselines, and the results are reported in Table~\ref{table:baseline_abla}.
We observe consistent improvements on all metrics by applying the proposal to PlaneRCNN~\cite{LiuKGFK19} and PlaneAE~\cite{YuZLZG19}.

\begin{table}
    \centering
    \begin{tabular}{c@{\hskip 6pt}c|c}
        \hline
        CoordConv~\cite{LiuLMSFSY18} & \textbf{yaw-invariant} & \makecell{V-plane orientation\\error (deg\degree)$\downarrow$} \\
        \hline
        \hline
        \multicolumn{3}{l}{\textbf{Stanford2D3D~\cite{ArmeniSZS17} validation set}}\\
        \hline
                   &            & 7.29 \\
        \checkmark &            & 5.97 \\
                   & \checkmark & \textbf{5.18} \\
        \checkmark & \checkmark & 5.21 \\
        \hline
    \end{tabular}
    \caption{
        We train four networks with different settings to compare CoordConv~\cite{LiuLMSFSY18} and the proposed yaw-invariant parameterization on V-plane orientation estimation.
    }
    \label{table:yawerr}
    \vspace{-1em}
\end{table}

\paragraph{Can the proposed plane orientation parameterization alone address the yaw ambiguity of \threesixty camera?}
To show the yaw ambiguity problem of \threesixty images and the proposed yaw-invariant representation can effectively address it, we train all combinations of the use of CoordConv and our yaw-invariant parameterization.
Table~\ref{table:yawerr} shows that the per-pixel normal error in degree over the V-planar pixels.
As mentioned in Sec.~\ref{ssec:geometry}, V-planar normals depend on \threesixty camera yaw rotation, but CNN layers cannot handle the yaw change very well.
Consequently, the model with neither CoordConv nor the proposed yaw-invariant parameterization yields inferior V-planar geometry quality.
By simply conditioning the model on image $u$-coordinates, the orientation error is decreased by a margin.
Finally, we show that re-parameterizing the plane parameter to be yaw-invariant can achieve superior orientation quality.
Employing CoordConv upon our parameterization does not improve the result, suggesting that the proposed parameterization alone has adequately solved the yaw ambiguity.

\begin{figure}
  \centering
  \includegraphics[width=\linewidth]{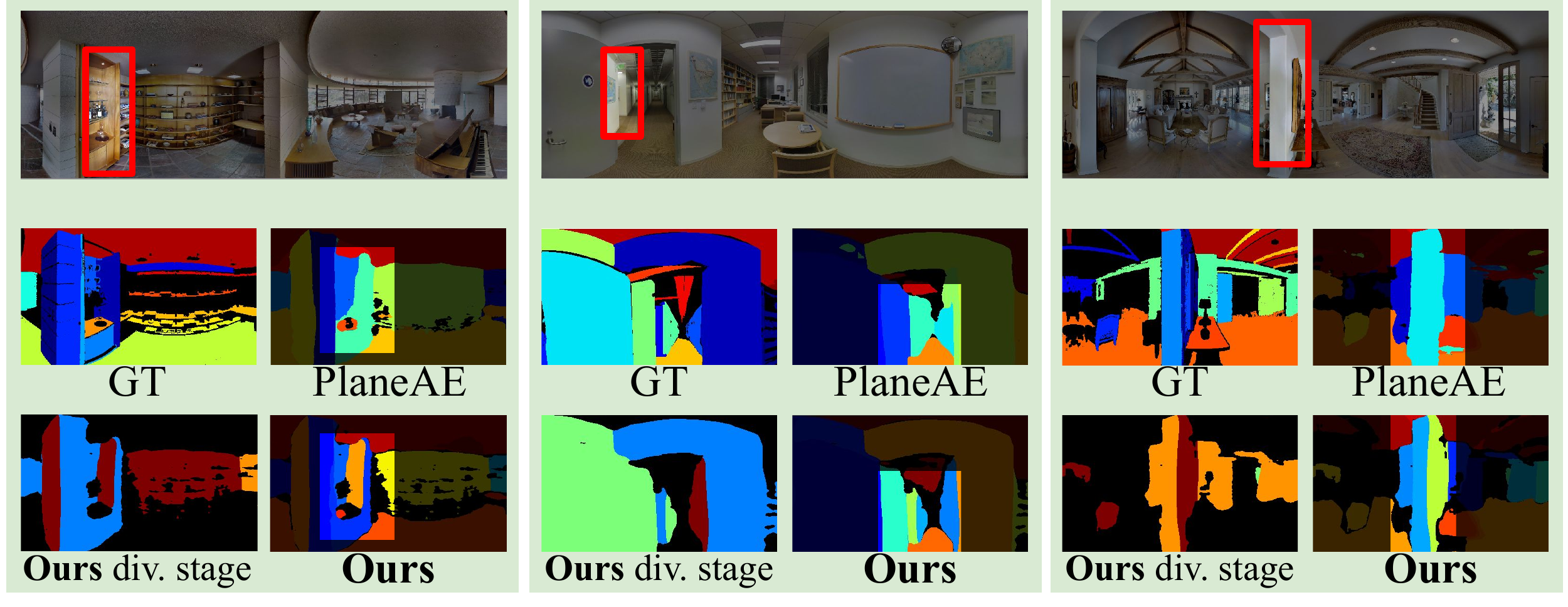}
  \caption{
    Three examples to demonstrate the effectiveness of our divide-and-conquer plane instance segmentation strategy.
    In comparison with PlaneAE~\cite{YuZLZG19}, which fails in the highlighted regions, our method can separate the undetected plane from its neighboring planes in the \emph{divide} stage.
  }
  \label{fig:qualpreseg}
  \vspace{-1.5em}
\end{figure}

\paragraph{The advantage of the \emph{divide} stage in our divide-and-conquer strategy.}
The \emph{divide} stage can early separate planes that are otherwise challenging to differentiate by the visual cues using the baseline method.
Three representative examples are highlighted in Fig.~\ref{fig:qualpreseg}, where we find some planes are merged by the baseline method PlaneAE~\cite{YuZLZG19} as it considers only pixel embedding. 
In contrast, these falsely merged planes can be separated early and easily by our \emph{divide} stage using orientation estimation.
%%%%%%%%% Experiment end

\section{Conclusion}
We propose a method that benefits from the  divide-and-conquer strategy of approximating an indoor panoramic scene by horizontal and vertical planes.
We point out a critical issue about the camera yaw ambiguity in panorama, which is neglected by previous methods.
A simple and effective yaw-invariant parameterization is proposed to solve the ambiguity adequately.
Last but not least, we construct the first real-world H\&V-planar reconstruction benchmark by extending the existing large-scale panorama datasets with H\&V-plane modality, and we apply two state-of-the-art plane instance segmentation methods to serve as the strong baselines for our dataset.
We show that our method achieves superior performance on this newly constructed benchmark.

\noindent \textbf{Acknowledgements:}
% This work was supported in part by the MOST, Taiwan under Grant 110-2634-F-001-009 and 110-2634-F-007-016, MOST Joint Research Center for AI Technology and All Vista Healthcare. We are grateful to the National Center for High-performance Computing for computer time and facilities.
This work was supported in part by the MOST, Taiwan under Grants 110-2634-F-001-009 and 110-2634-F-007-016, MOST Joint Research Center for AI Technology and All Vista Healthcare. We thank National Center for High-performance Computing (NCHC) for providing computational and storage resources.

%%%%%%%%%%%% Supplementary material
\clearpage
% \onecolumn
\setcounter{section}{0}
\renewcommand\thesection{\Alph{section}}
\noindent{\bf\Large Supplementary material}

\section{Ground-truth H\&V-plane annotation} \label{sec:gt}
This section provides more details about the extraction of \emph{HV-planes} from ground-truth depth.

\subsection{Local analysis}
The goal is to classify each pixel into `H-pixel', `V-pixel', or `other' for later HV-plane extraction.
The same rule is generally applied to all pixels, so we describe how we classify a pixel $p$ for simplicity as follows.
We relocate the world origin to the 3D coordinate of point $p$ (same process applied to other points); the z-axis is aligned with gravity, and all the 3D points from the same image column are on the plane formed by y-axis and z-axis.
Let $(0, y_t, z_t)$ and $(0, y_b, z_b)$ be the 3D coordinates from the top and bottom adjacent pixels respectively.
Let $d$ be the depth value of pixel $p$.
We calculate the following information:
\begin{algorithmic}
    \STATE $T_H \gets \min(d \cdot 5, ~ 40)$  \COMMENT{threshold for `H'}
    \STATE $T_V \gets 90 - T_H$  \COMMENT{threshold for `V'}
    \STATE $\theta_t \gets \arctantwo(|z_t|, |y_t|)$  \COMMENT{degree in range $[0^{\circ}, 90^{\circ}]$}
    \STATE $\theta_b \gets \arctantwo(|z_b|, |y_b|)$  \COMMENT{degree in range $[0^{\circ}, 90^{\circ}]$}
    \STATE $type_t \gets \text{thresholdHVO}(\theta_t, T_H, T_V)$
    \STATE $type_b \gets \text{thresholdHVO}(\theta_b, T_H, T_V)$
\end{algorithmic}
\noindent $T_H$ and $T_V$ are depth-dependent thresholds (the farther, the more tolerant). The procedure `thresholdHVO' is defined as 

\begin{algorithmic}
    \IF {$\theta < T_H$ \AND $\max(|z_t|, |z_b|) < 0.1$}
        \STATE $type \gets \text{`H'}$
    \ELSIF {$\theta > T_V$ \AND $\max(|y_t|, |y_b|) < 0.1$}
        \STATE $type \gets \text{`V'}$
    \ELSE
        \STATE $type \gets \text{`O'}$
    \ENDIF
\end{algorithmic}

\noindent We then classify the pixel based on the states of $type_t$ and $type_b$ using the rule depicted in Table~\ref{table:firstrule}.
The `TBD' in Table~\ref{table:firstrule} needs to be further determined by $\theta_t$ and $\theta_b$:
\begin{algorithmic}
    \IF {$type_t = \text{`V'}$ \AND $type_b = \text{`H'}$}
        \IF {$90 - \theta_t < \theta_b$}
            \RETURN `V-pixel'
        \ELSE
            \RETURN `H-pixel'
        \ENDIF
    \ELSIF {$type_t = \text{`H'}$ \AND $type_b = \text{`V'}$}
        \IF {$\theta_t < 90 - \theta_b$}
            \RETURN `H-pixel'
        \ELSE
            \RETURN `V-pixel'
        \ENDIF
    \ENDIF
\end{algorithmic}

\begin{table}[h]
    \centering
    \begin{tabular}{c|c|c|c}
        $type_t ~/~ type_b$ & `H' & `V' & `O' \\
        \hline
        `H' & `H-pixel' & TBD & `H-pixel' \\
        \hline
        `V' & TBD & `V-pixel' & `V-pixel' \\
        \hline
        `O' & `H-pixel' & `V-pixel' & `other' \\
    \end{tabular}
    \caption{
    Type of a pixel according to the relationship with its top and bottom adjacent pixels
    }
    \label{table:firstrule}
\end{table}

\noindent Finally, we use connected component (CC) analysis on `H-pixel' and `V-pixel', and re-assign small CC into `other'.

\subsection{Horizontal planes extraction}
A horizontal plane has only one degree of freedom that can be parameterized by the signed distance (for up/down in 3D) to the camera height.
We identify horizontal planes in a manner similar to non-maximum suppression:
\begin{enumerate}[itemsep=0pt]
    \item Using linear search to find an H-plane covering the largest number of remaining H-pixels within 5cm;
    \item Removing any connected component (CC) on the image that contains fewer than 1{,}000 pixels (\ie $\approx 0.2\%$ of the image size);
    \item Masking out inlier H-pixels.
\end{enumerate}
The process iterates until no H-plane can be found.
The extracted H-planes are refined by reassigning all H-pixels to their nearest H-planes and recalculating the parameters of H-planes based on new inlier H-pixels.

\subsection{Vertical planes extraction}
A vertical plane has two degrees of freedom, which can be characterized by $\vec{n} = [x\; y \;0] = o \cdot [\cos\theta \; \sin\theta \; 0]$---the unit surface normal $[\cos\theta \; \sin\theta \; 0]$ scaled by the plane offset $o$.
Similar to H-plane extraction, we apply the following process:
\begin{enumerate}[itemsep=0pt]
    \item RANSAC finding a plane $\vec{n}$ that covers the largest number of V-pixels within a threshold $T = \min(o\cdot 0.05,\, 0.2)$;
    A 3D point is said to be an inlier if the distance to the plane is less than the threshold;
    \item Keeping only the largest CC as a V-plane instance since the detected plane $\vec{n}$ by RANSAC extends infinitely in 3D and could cover multiple non-connected planes.
    \item Masking out inlier V-pixels.
\end{enumerate}
The process iterates until the largest CC contains fewer than 1{,}000 pixels.
The extracted V-planes are refined by reassigning border pixels of instances to their nearest V-planes.
This step could cut an instance into different CCs; only the largest CC of the instance remains, and the other smaller-sized CCs are reassigned to the majority of their neighbors.
Finally, V-plane parameters are refined by solving the least-squares based on the new inliers.

\section{Relation with panorama room layout}
The room layout reconstruct the highest-level structure of a room, which is consists of only a few facades and the result is up to a scale.
Besides, most of the room layout reconstruction methods currently only consider one-floor-one-ceiling model.
In plane evaluation metrics, many factors can even makes the detected layout facades to be counted as false positive, \eg, planes of furniture or objects taking a large amount of pixels of a wall, beam, column, gates or doors to another area.
% On the other hand, in our plane detection datasets, there are roughly 20 planes from a panorama in average.

% For large objects/furniture or 
% Besides, planes 
% some large objects or furniture (\eg, table) are consider as a part of floor or wall in room layout, which hurts the evaluation metric for plane.

\section{Detailed network architecture}
\begin{figure}
    \centering
    \includegraphics[width=.8\linewidth]{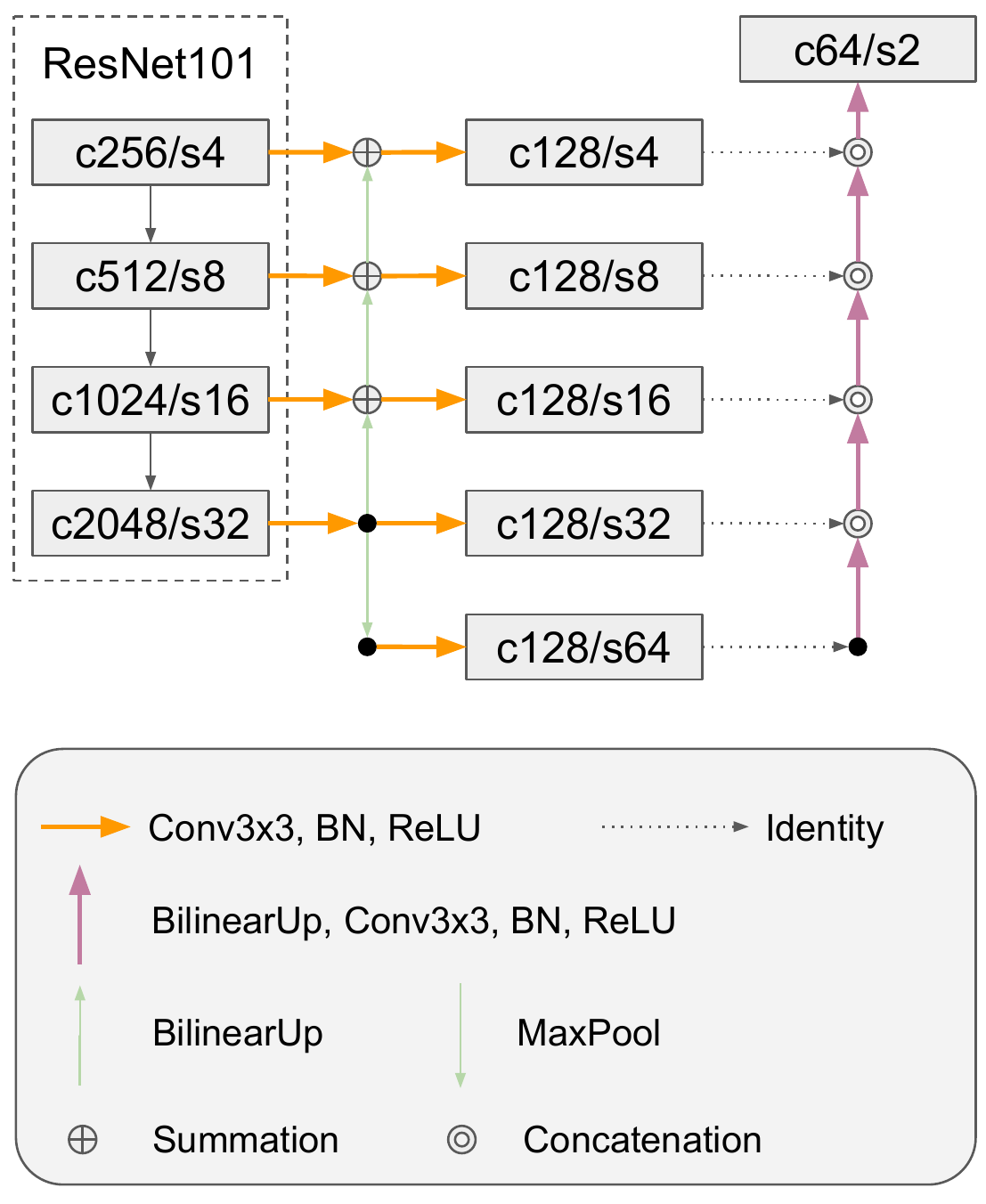}
    \caption{
        Detailed network architecture employed to extract per-pixel features.
        The ``c$X$/s$Y$'' indicates that the feature tensor has $X$ latent channels with spatial stride $Y$.
    }
    \label{fig:arch_detail}
\end{figure}

The architecture for pixel-level feature extraction is depicted in Fig.~\ref{fig:arch_detail}.
We employ ResNet101 as our backbone network, which provides features in four different output stride.
We use $\operatorname{ConvBlock}$ to denote a sequence of $\operatorname{Conv3x3}, \operatorname{BN}, \operatorname{ReLU}$.
Each backbone feature tensor is reduced to $128$ latent channels by two $\operatorname{ConvBlocks}$, where the intermediate features are summed with the upsampled coarser-scale features to capture a longer range of context.
We also adding one extra level by $\operatorname{MaxPool}$.
Finally, from coarse to fine levels, the features are upsampled, refine by $\operatorname{ConvBlock}$ and concatenated with the finer level features.
The overall encoder-decoder network produce a pixel-level feature tensor at output stride $2$ with $64$ latent channels, which is then followed by different $\operatorname{Conv1x1}$ layer to estimate each modality for HV-planes reconstruction.

\section{Visualization of yaw-rotation ambiguity in \threesixty images} \label{sec:yaw}
\begin{figure}[h]
    \centering
    \begin{tabular}{@{}c@{\hskip 10pt}c@{}}
        \includegraphics[width=0.4\linewidth]{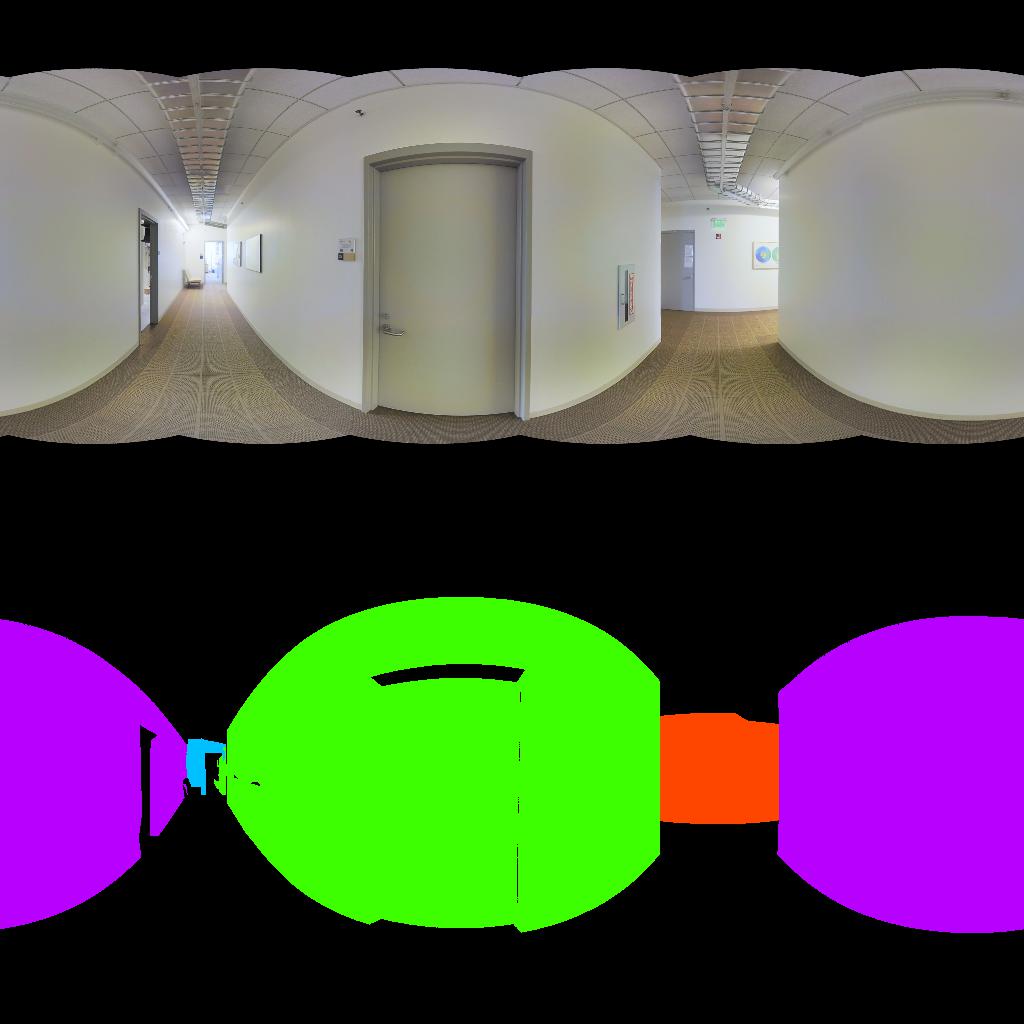} & \includegraphics[width=0.4\linewidth]{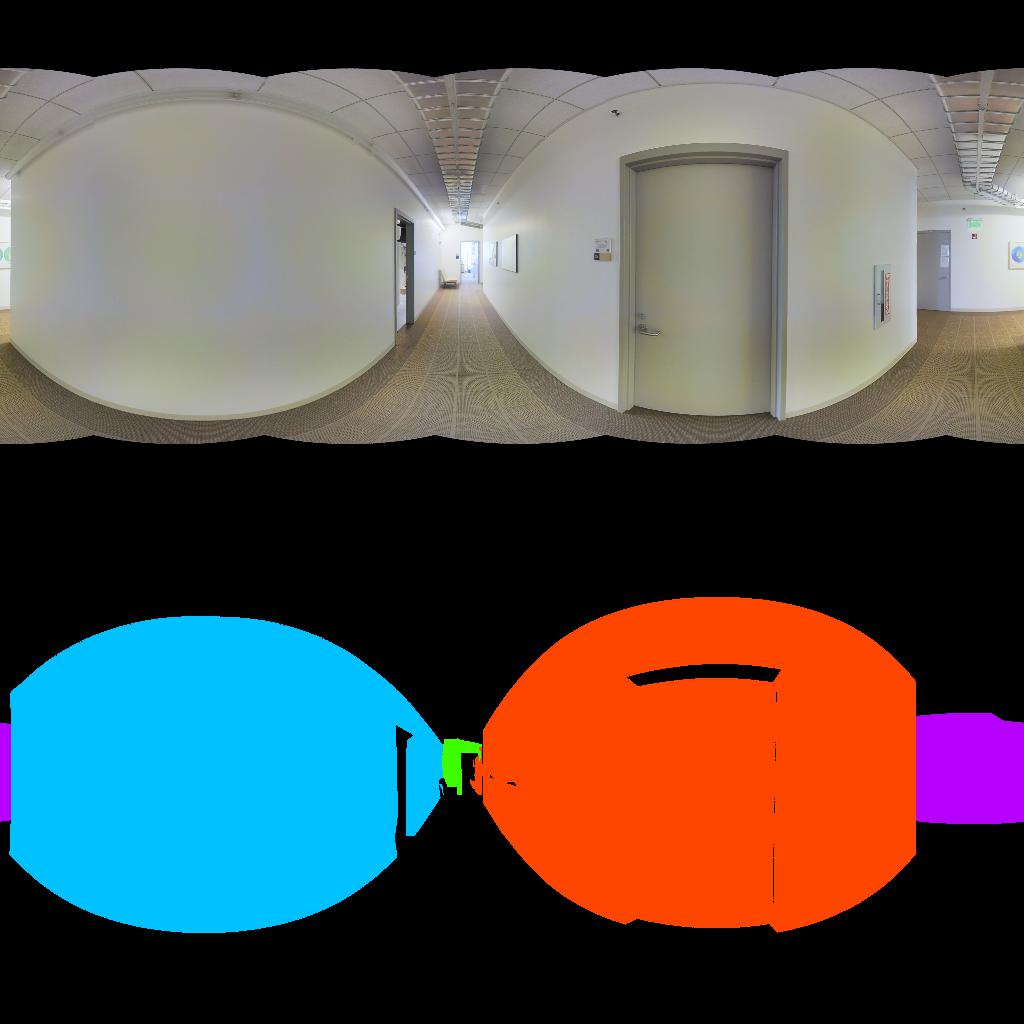}
    \end{tabular}
    \caption{
        RGB and ground-truth V-planes colored by the angles of plane orientations.
        The left and right figures are two views of the exact same scene with different camera yaw rotations.
    }
    \label{fig:yaw}
\end{figure}

We visualize the effect of camera yaw rotation on V-plane parameters in a \threesixty image in Fig~\ref{fig:yaw}, where planes are colored according to the angles of the plane orientations.
We can see that values of the standard plane representation vary with different yaw rotations of \threesixty camera. This property, however, is inconsistent with the fact that the convolutional neural network is translation-invariant and therefore less aware of the yaw rotation.

\begin{figure}[h]
    \centering
    \begin{tabular}{@{}c@{\hskip 10pt}c@{}}
        \includegraphics[width=0.4\linewidth]{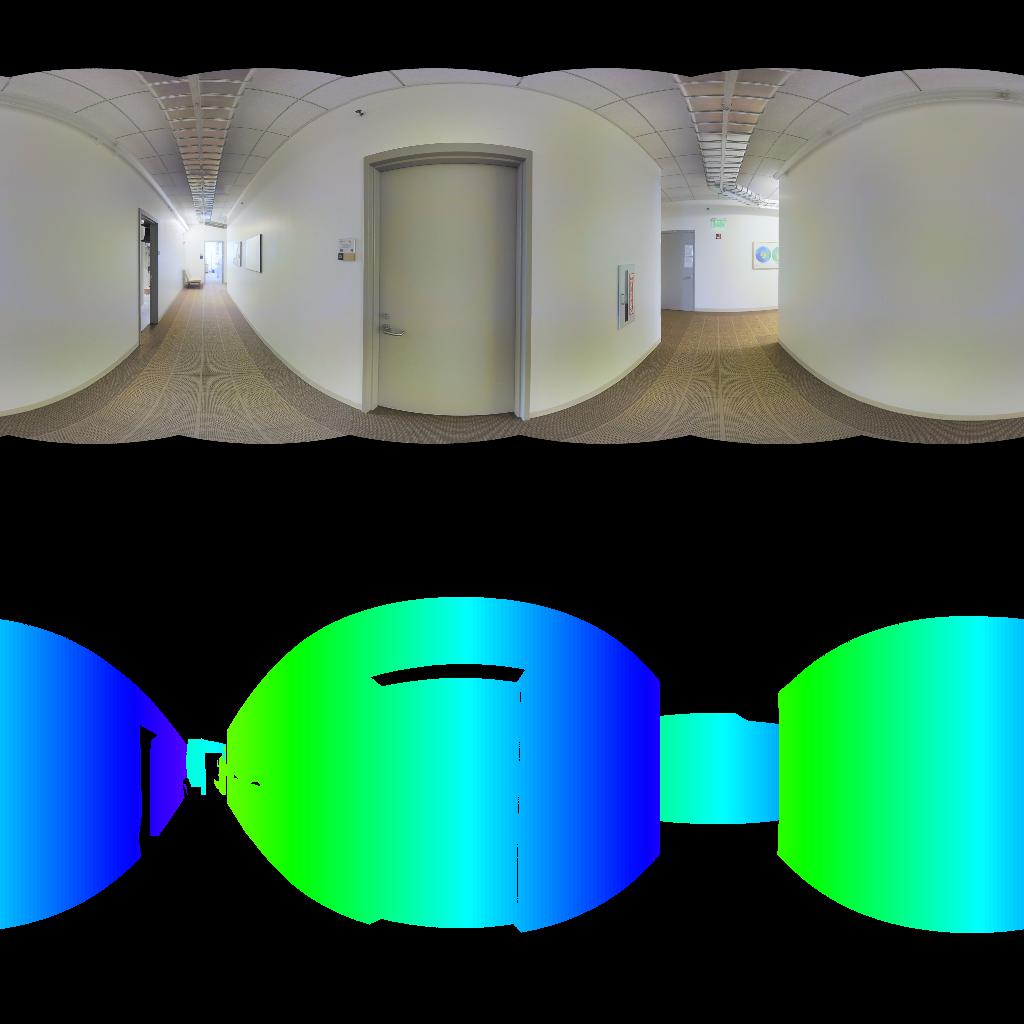} & \includegraphics[width=0.4\linewidth]{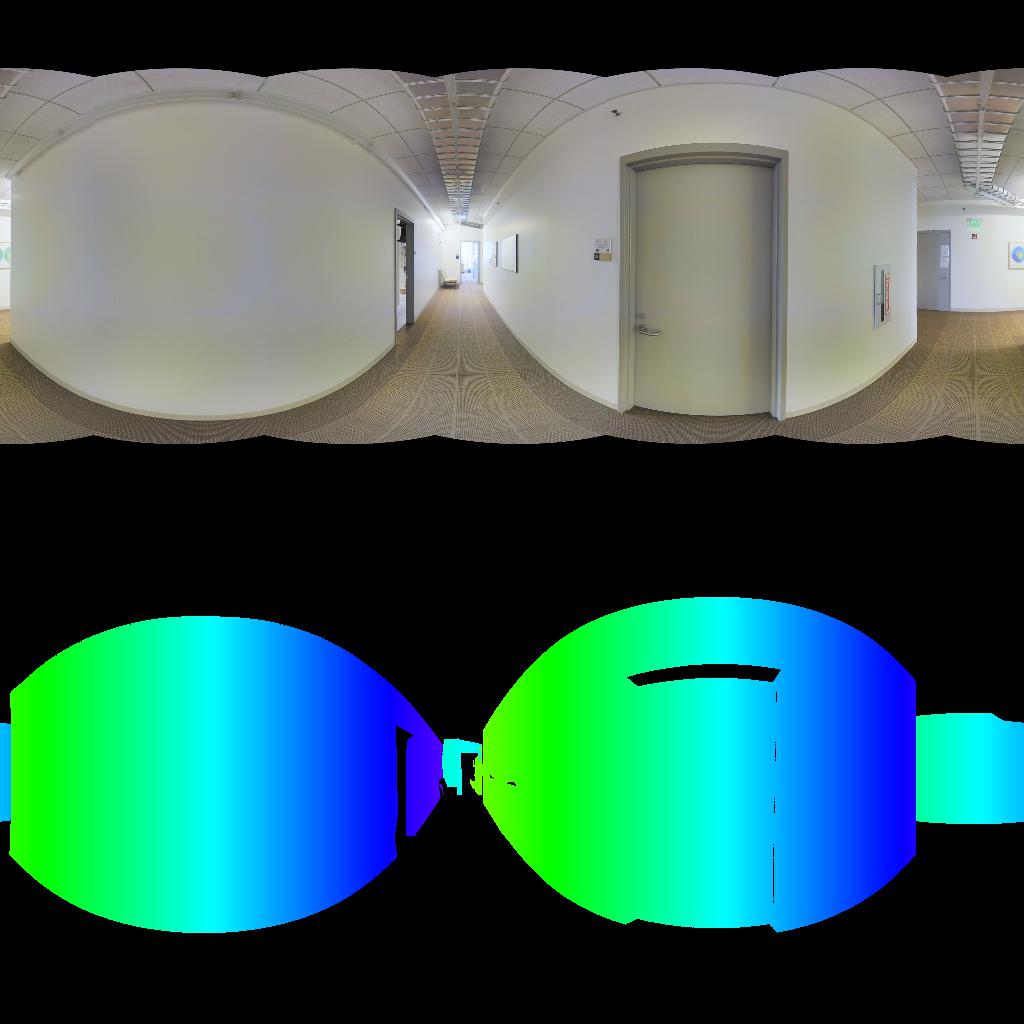}
    \end{tabular}

    \caption{
        The proposed camera yaw-invariant plane representation.
    }
    \label{fig:yaw_invariant}
\end{figure}

To address the yaw ambiguity effectively, we propose to re-parameterize the ground-truth plane orientation $\theta_i^*$ of each pixel $i$ into \emph{residual form} with respect to the pixel yaw viewing angle $u_i$ such that it is invariant to the \threesixty camera yaw rotation:
\begin{equation}
{\theta'}_i^* = \theta_i^* - u_i \,.
\end{equation}
We show the same scene as Fig~\ref{fig:yaw} but with our camera yaw invariant plane representation in Fig~\ref{fig:yaw_invariant}, where the ground-truth orientation is now camera yaw-invariant.

\section{V-planar orientation grouping}
Our proposed plane instance segmentation module is geometry-aware.
More specifically, we group all pixels with similar plane orientations in the first stage of the plane instance segmentation process.
We describe below how we instantiate the grouping process for pixels classified as `V-planar'.
Let $\theta_i \in [-\pi, \pi]$ denote the reconstructed V-plane orientation in radian of a V-planar pixel with index $i$.
We discretize all $\{\theta_i \mid \text{pixel } i \text{ is V-planar}\}$ into 360 bins.
Note that the histogram is circular, where the first bin is adjacent to the last bin.
We then find all prominent peaks on the histogram with two criteria: \textit{i}) the vote is larger than any bins within the nearest 50 bins, and \textit{ii}) it has at least 100 votes.
V-pixels are assigned to their nearest peak to form surface orientation groups.

% An example of the voting result is shown in Fig.~\ref{fig:preseg}.

% \begin{figure}
%     \centering
%     \includegraphics[width=0.27\linewidth]{fig/preseg.pdf}

%     \caption{
%         One result of surface orientation grouping.
%         Different colors denote different groups.
%         The outer circle represents the distribution of yaw-rotations, where each point on the circle is a vote by a V-planar pixel.
%         This is the same figure as shown in main paper Fig.~5b
%     }
%     \label{fig:preseg}
% \end{figure}

\section{More quantitative comparison of intermediate version}
We show more intermediate version of our method in Table~\ref{table:more_abla}.
Our based method (w/o {\it yaw-inv}, w/o {\it ori-gp}) shows slightly better result on pixel-recall and plane-recall comparing to PlaneAE, which could be due to the difference in model architecture and the output modalities for representing plane parameter.
By applying the {\it ori-gp}, we can see consistent improvement on segmentation quality, especially when it works with the {\it yaw-inv} (the quality of {\it ori-gp} depend on the quality of the planar yaw-angle estimation).
Further, adding {\it yaw-inv} to our system leads to a significant improvement on pixel-recall and plane-recall which considering not only the segmentation quality but also geometry quality.
\begin{table}
    \centering
    \begin{tabular}{l|ccc|cc}
        \hline
        \multirow{2}{*}{Method} &
        \multicolumn{3}{c|}{Segmentation Quality} &
        \multicolumn{2}{c}{Recall$\uparrow$}\\
         & ARI$\uparrow$ & VI$\downarrow$ & SC$\uparrow$ & Pixel & Plane \\
        \hline
        PlaneAE~\cite{YuZLZG19} & 0.765 & 1.536 & 0.733 & 0.520 & 0.341\\
        \hline\hline
        \makecell[l]{w/o {\it yaw-inv}\\w/o {\it ori-gp}} & 0.762 & 1.555 & 0.734 & 0.542 & 0.360\\
        \hline
        \makecell{w/o {\it yaw-inv}} & 0.764 & 1.542 & 0.738 & 0.545 & 0.369\\
        \hline
        w/o {\it ori-gp} & 0.762 & 1.554 & 0.732 & 0.619 & 0.417\\
        \hline
        full & 0.768 & 1.514 & 0.742 & 0.627 & 0.430\\
        \hline
    \end{tabular}
    \caption{
        More quantitative results on Stanford2d3d of our method's intermediate version.
        A strong baseline---PlaneAE---is also shown for comparison.
        {\it yaw-inv}: the yaw-invariant representation.
        {\it ori-gp}: the orientation grouping as a pre-filtering before mean shift.
    }
    \label{table:more_abla}
    \vspace{-2em}
\end{table}

\section{More qualitative results reconstructed by our method.}
We show more visual results on the two real-world dataset---Stanford2D3D~\cite{ArmeniSZS17} dataset and Matterport3D~\cite{ChangDFHNSSZZ17} dataset---in Fig.~\ref{fig:more_quals_2d3d}, Fig.~\ref{fig:more_quals_mp3d} and Fig.~\ref{fig:more_qual_mesh}.

% In both figures, input RGB panoramas are shown in first column; the 2D plane instance segmentations detected by the proposed YawP$^3$ are shown in second column; the planar depth absolute difference with ground truth is shown in third column where the blue and red color represent error ranging from 0 to 3 meters, and the snapshots in the reconstructed 3D are show in last column.
% The images are stretched a little to fit in paper.

\section{Qualitative comparisons with baselines}
\begin{figure*}
    \centering
    \includegraphics[width=0.82\linewidth]{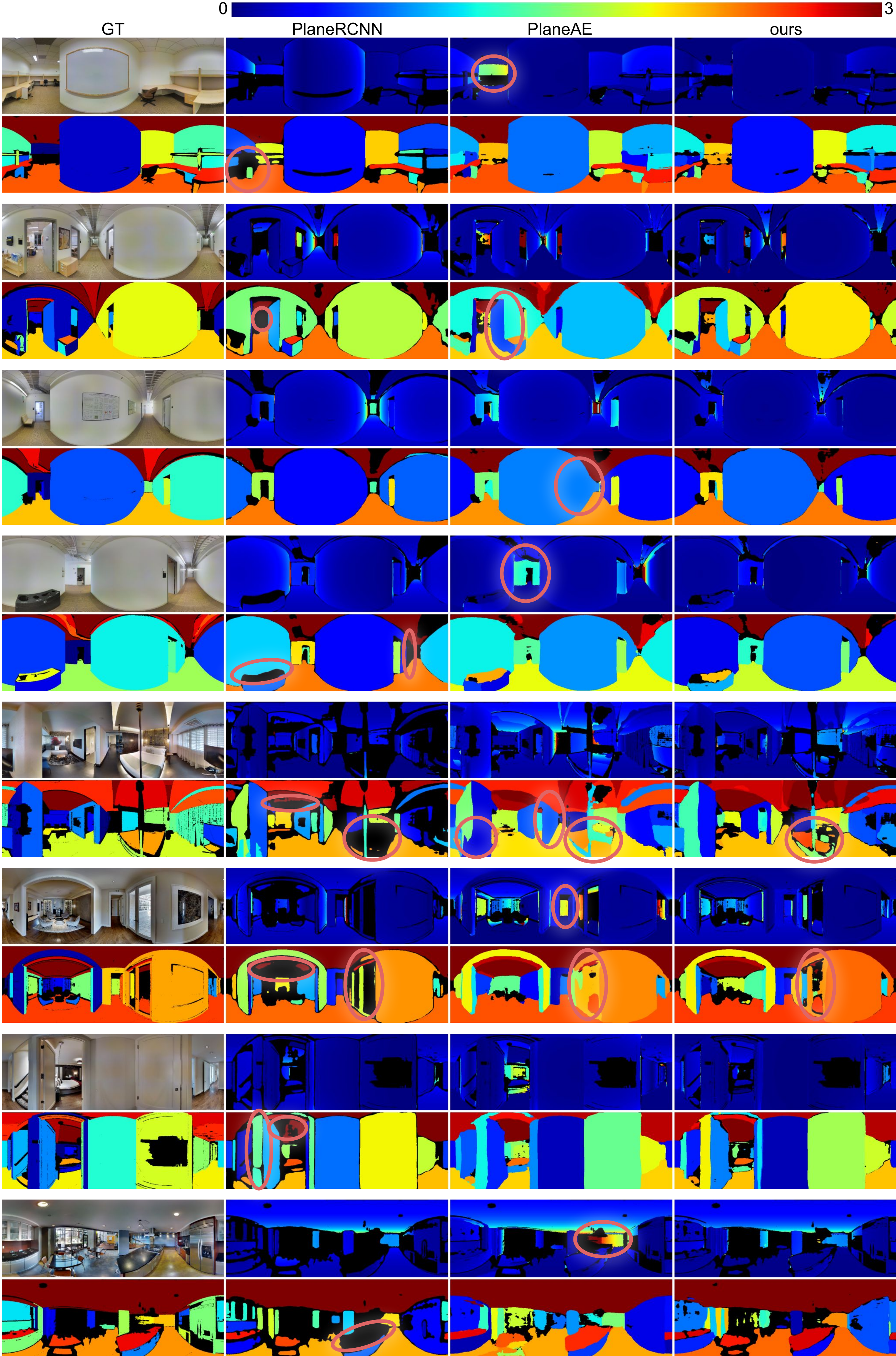}
    \caption{
        Qualitative comparisons with baselines. (See main text for details.)
    }
    \label{fig:more_quals_2d}
\end{figure*}

We compare more qualitative results of the two baselines---PlaneRCNN~\cite{LiuKGFK19} and PlaneAE~\cite{YuZLZG19}---and our method in Fig.~\ref{fig:more_quals_2d}.
The first column shows input RGB and ground-truth plane instance segmentation.
The second to the fourth columns respectively show results reconstructed by PlaneRCNN~\cite{LiuKGFK19}, PlanarReconstruct~\cite{YuZLZG19}, and our method.
For each scene, we show the planar depth error (clipped to 3 meters) in the top row and the predicted plane instance segmentation in the bottom row.
% For each example, the first row is the planar depth error (clipped to 3 meters), and the second row is the predicted plane instance segmentation.
We use red circles to highlight obvious errors.

\begin{figure*}
    \centering
    \includegraphics[width=0.95\linewidth,height=0.43\linewidth]{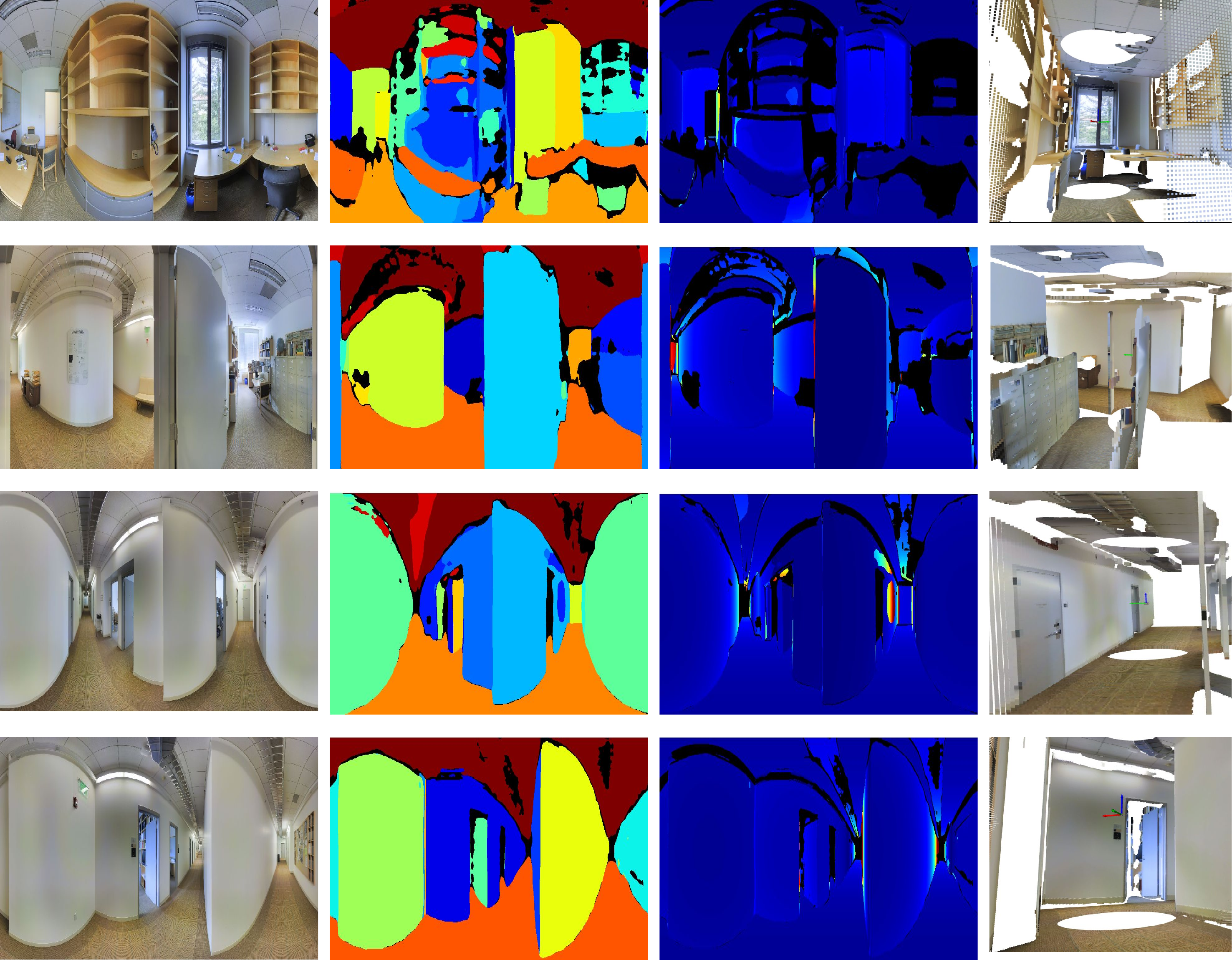}
    \caption{
        Visual results on Stanford2D3D~\cite{ArmeniSZS17} validation set.
        The first to the fourth columns show input RGB, plane instance segmentation detected by our method, planar depth error, and snapshot in the reconstructed 3D planar model.
    }
    \label{fig:more_quals_2d3d}
\end{figure*}

\begin{figure*}
    \centering
    \includegraphics[width=0.95\linewidth,height=0.65\linewidth]{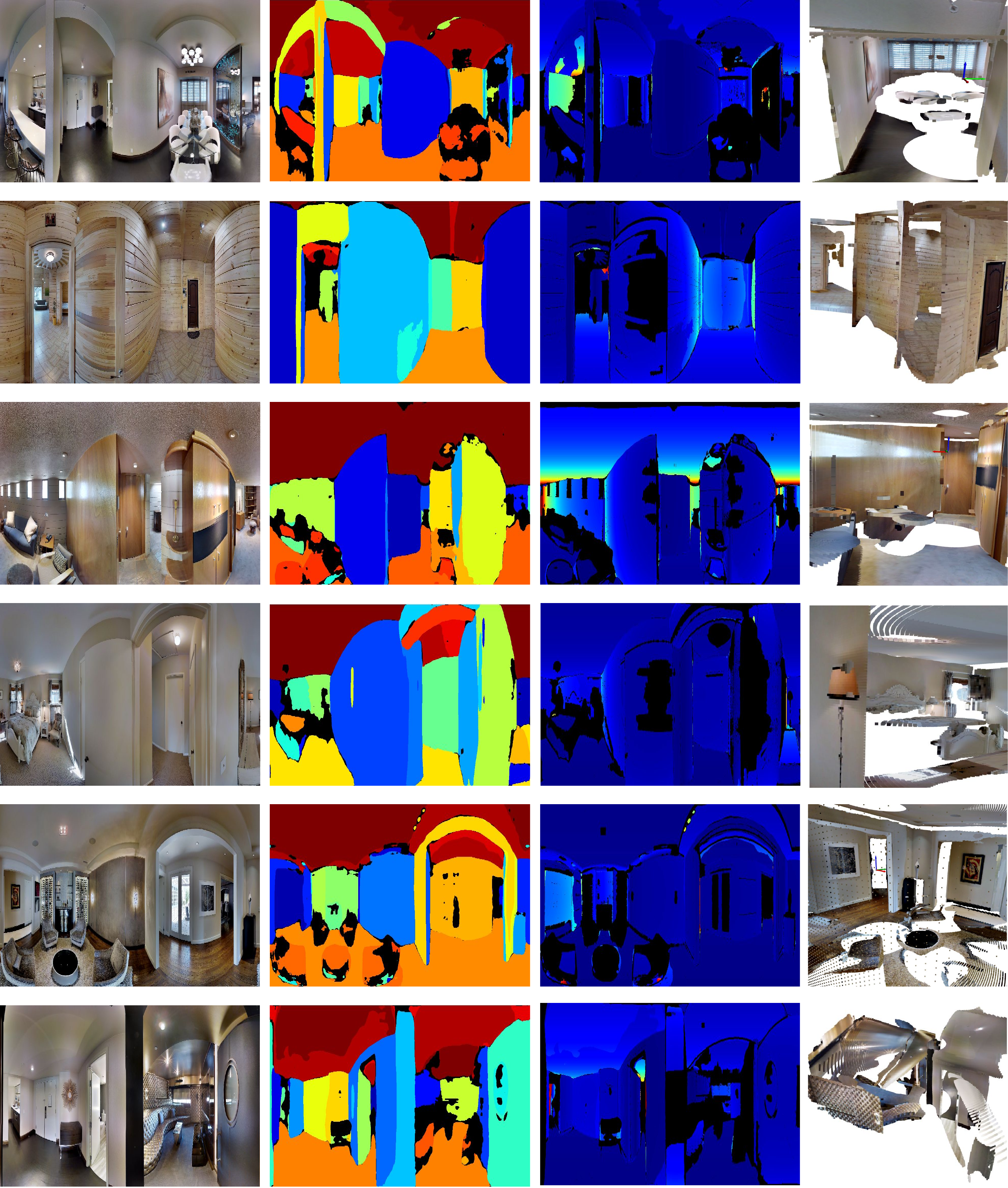}
    \caption{
        Visual results on Matterport3D~\cite{ChangDFHNSSZZ17} test set.
        The first to the fourth columns show input RGB, plane instance segmentation detected by our method, planar depth error, and snapshot in the reconstructed 3D planar model.
    }
    \label{fig:more_quals_mp3d}
\end{figure*}

\begin{figure*}
    \centering
    \begin{tabular}{@{}c@{}c@{\hskip 10pt}c@{}c@{}}
        \includegraphics[width=0.23\linewidth]{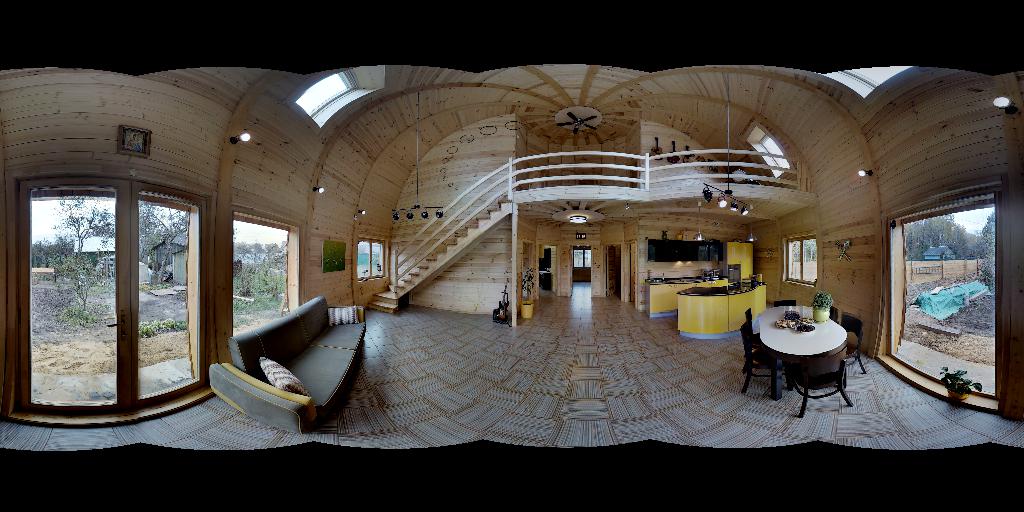} & \includegraphics[width=0.23\linewidth]{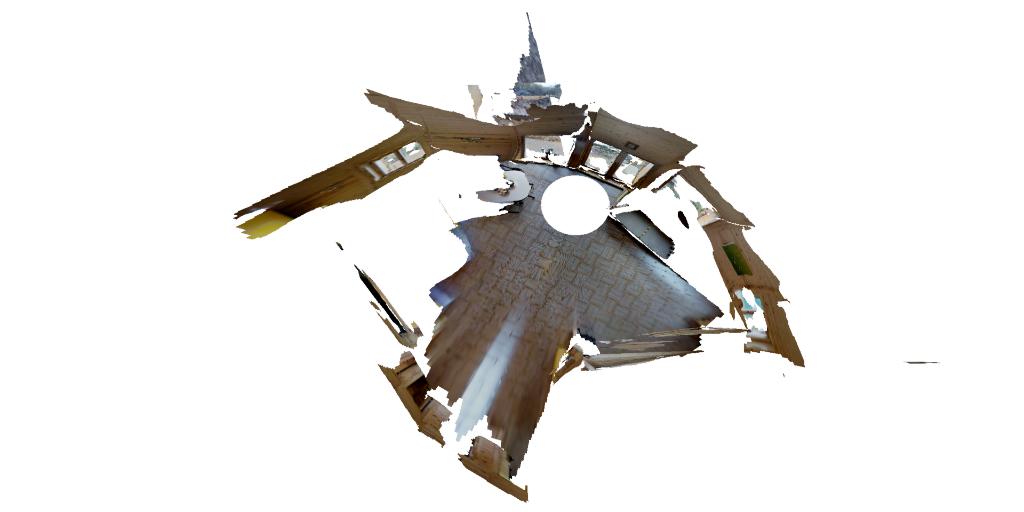} &
        \includegraphics[width=0.23\linewidth]{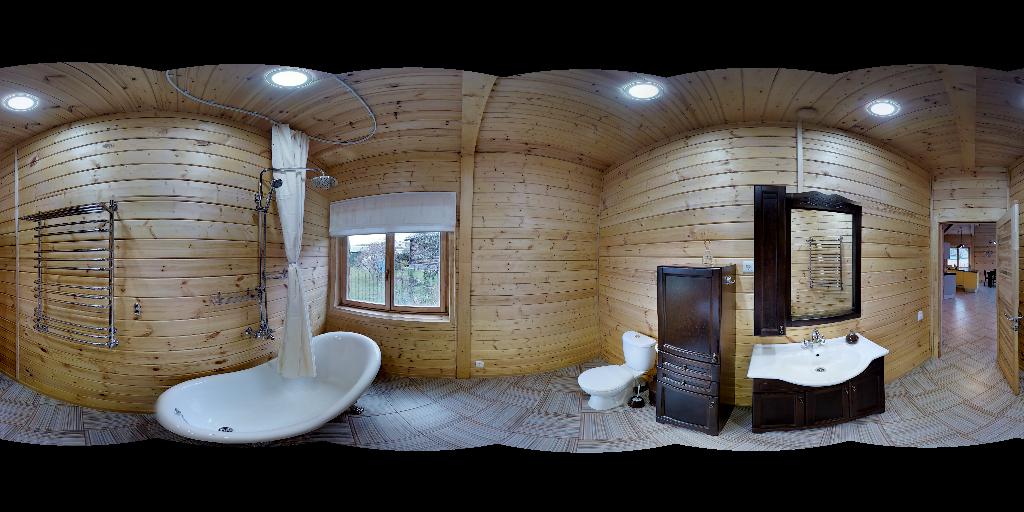} & \includegraphics[width=0.23\linewidth]{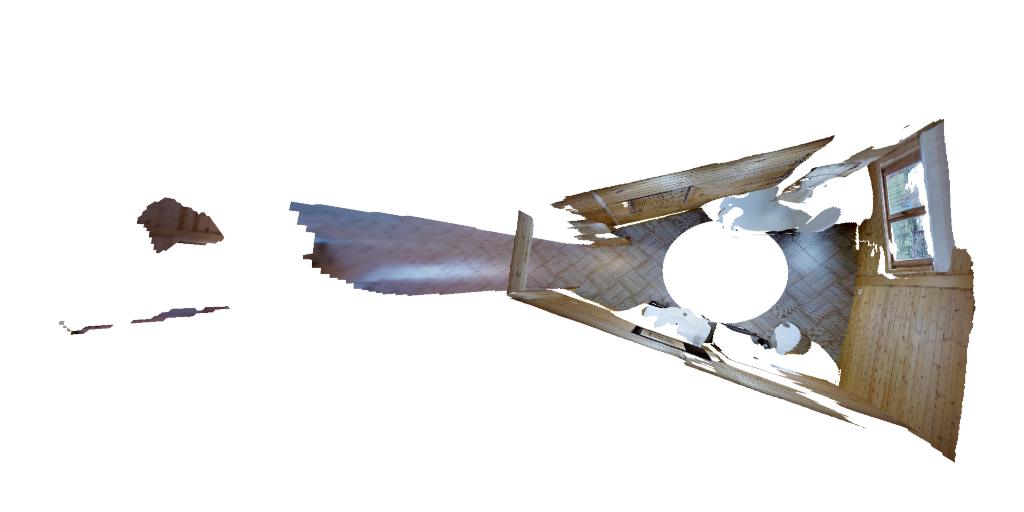} \\
        \includegraphics[width=0.23\linewidth]{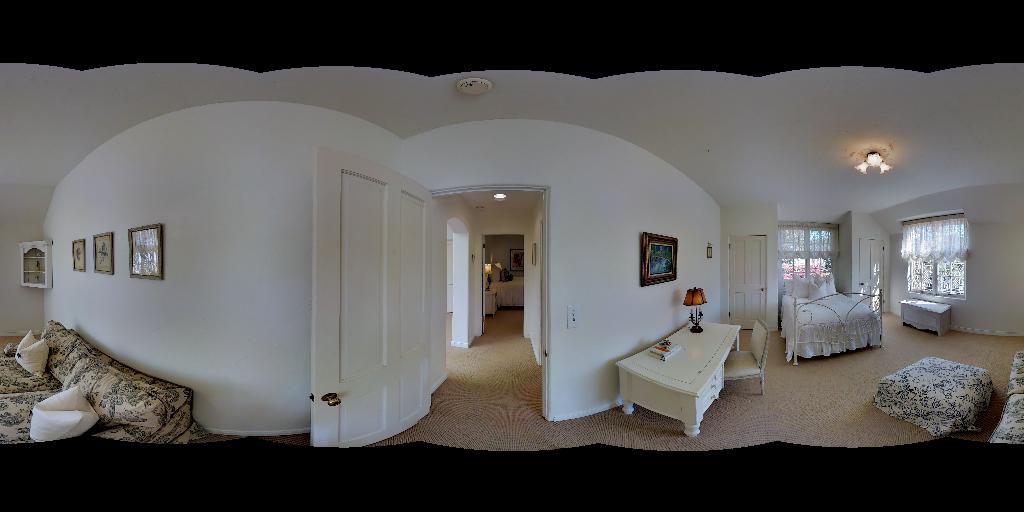} & \includegraphics[width=0.23\linewidth]{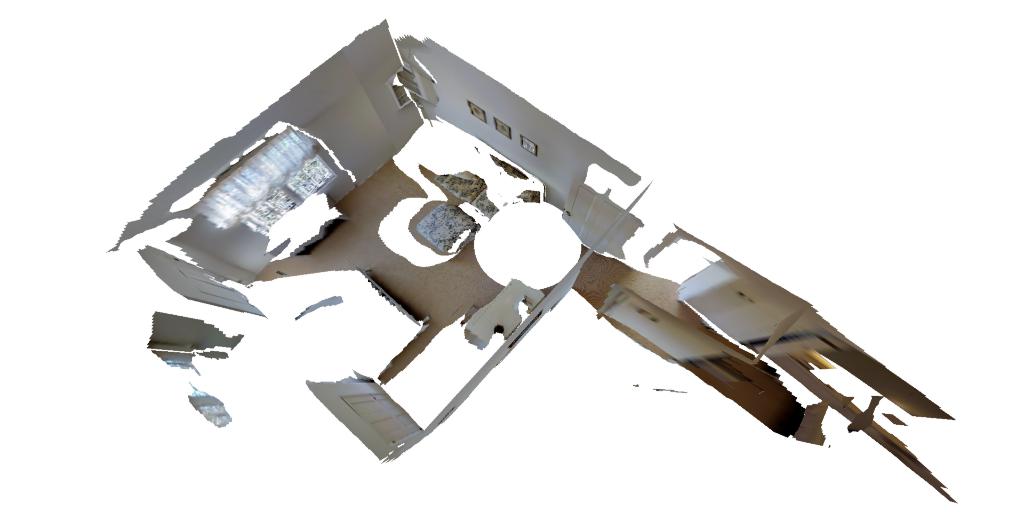} &
        \includegraphics[width=0.23\linewidth]{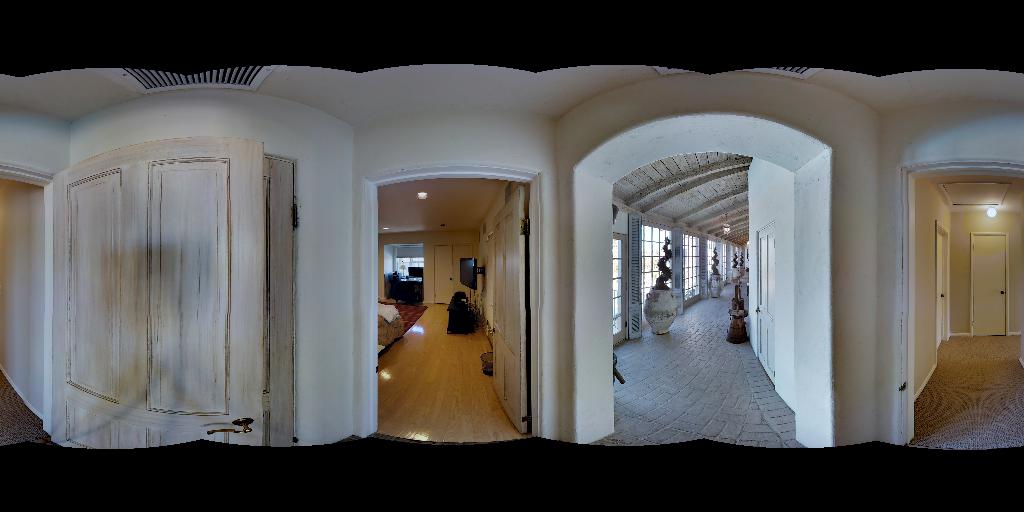} & \includegraphics[width=0.23\linewidth]{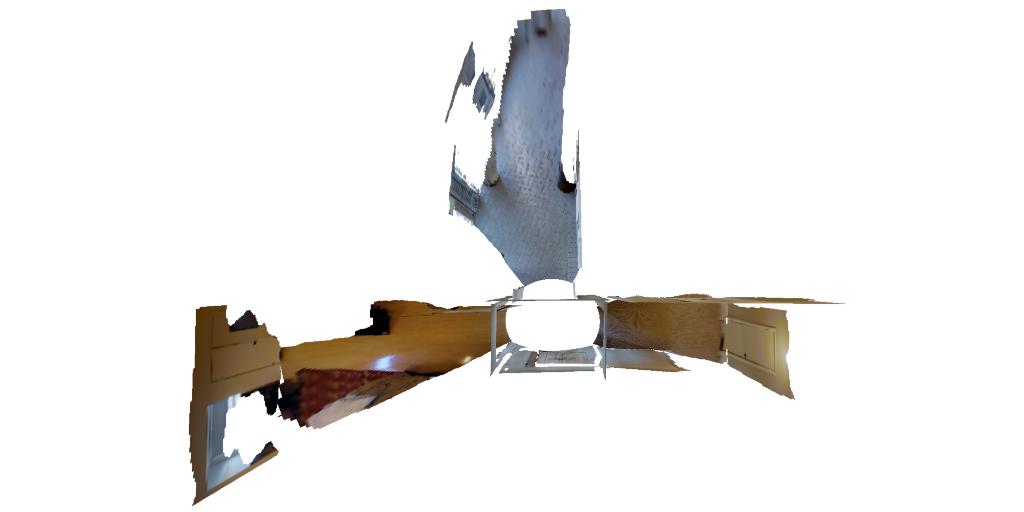} \\
        \includegraphics[width=0.23\linewidth]{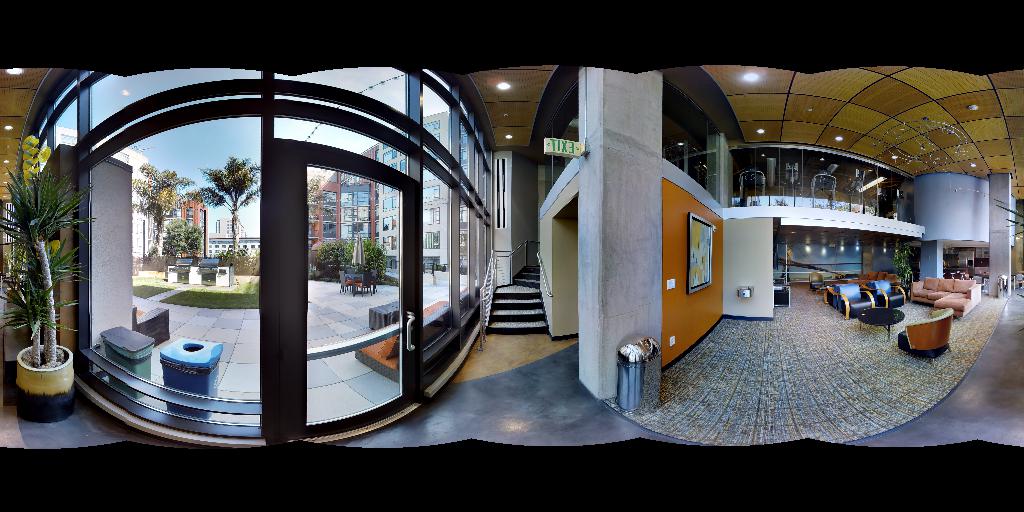} & \includegraphics[width=0.23\linewidth]{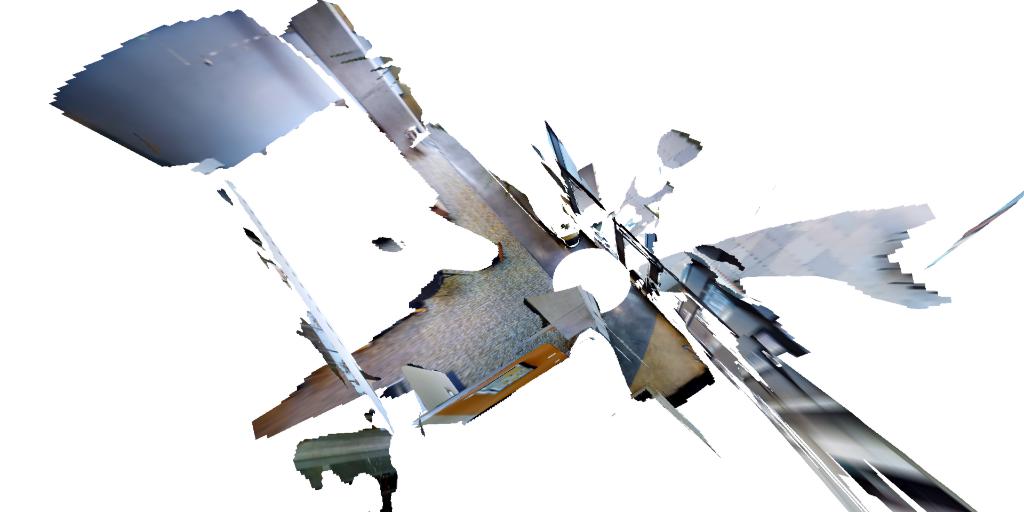} &
        \includegraphics[width=0.23\linewidth]{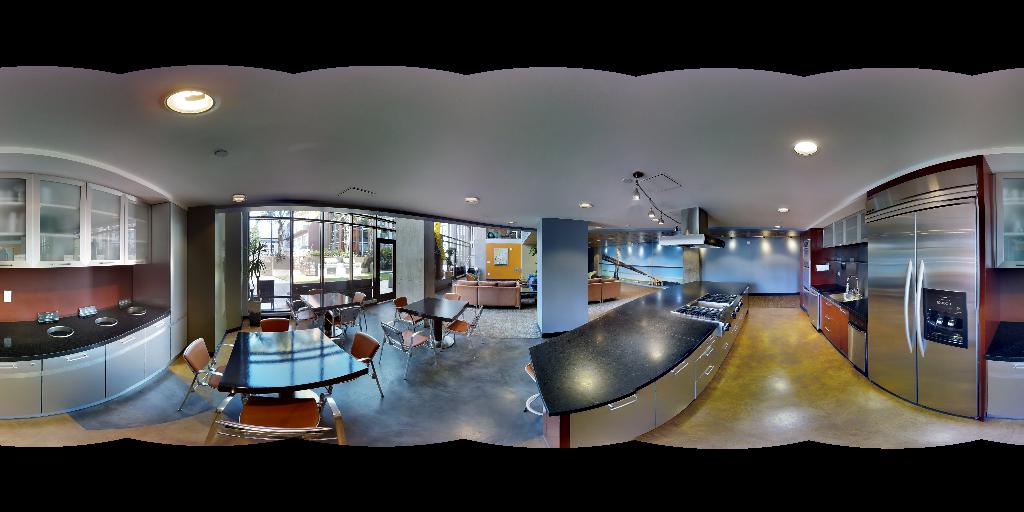} & \includegraphics[width=0.23\linewidth]{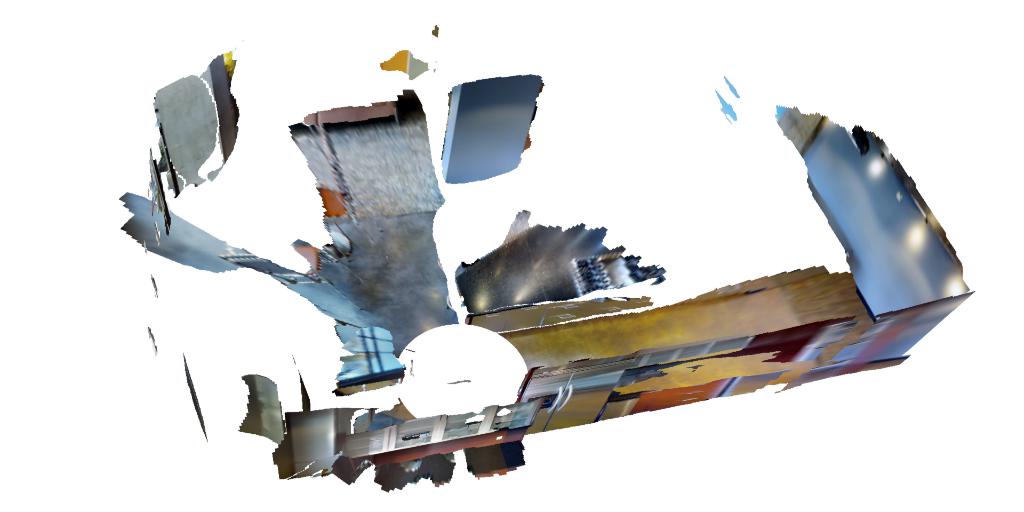} \\
        \includegraphics[width=0.23\linewidth]{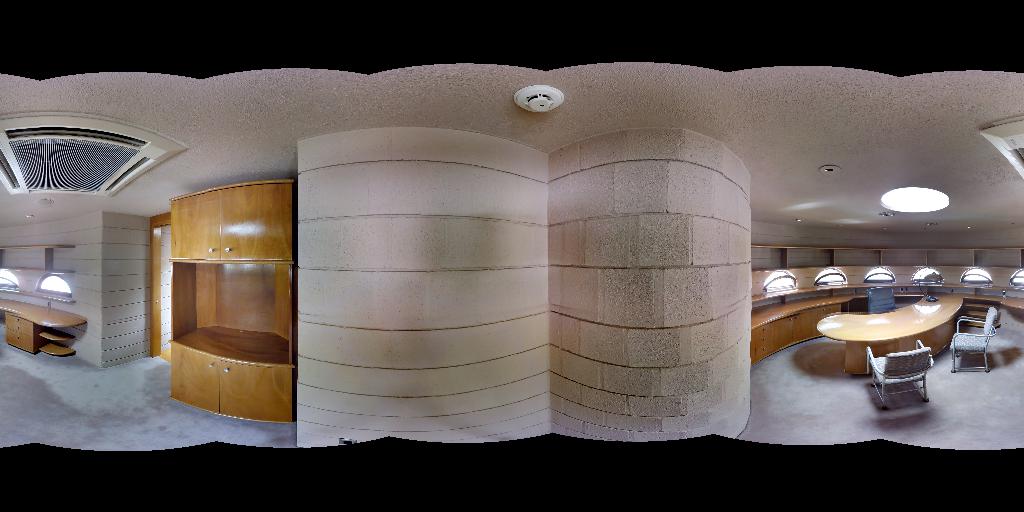} & \includegraphics[width=0.23\linewidth]{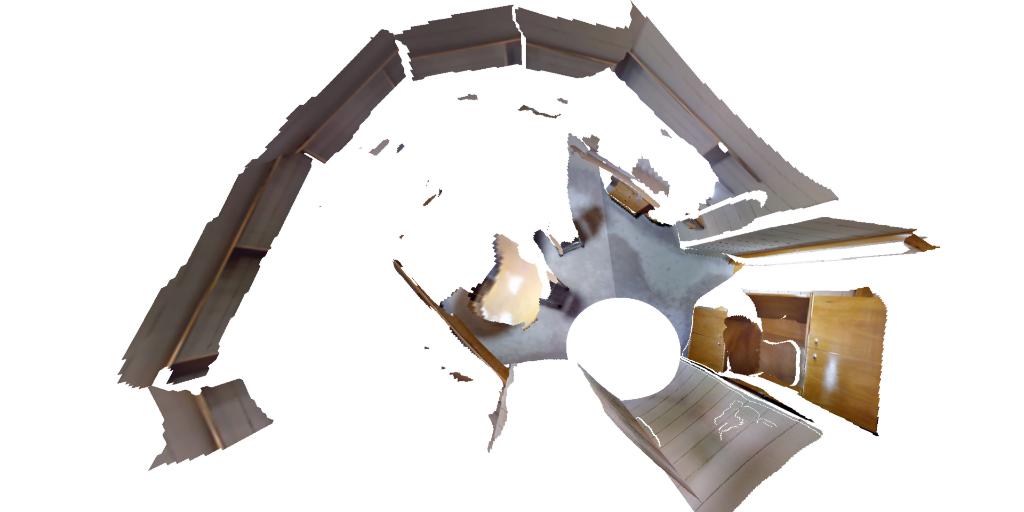} &
        \includegraphics[width=0.23\linewidth]{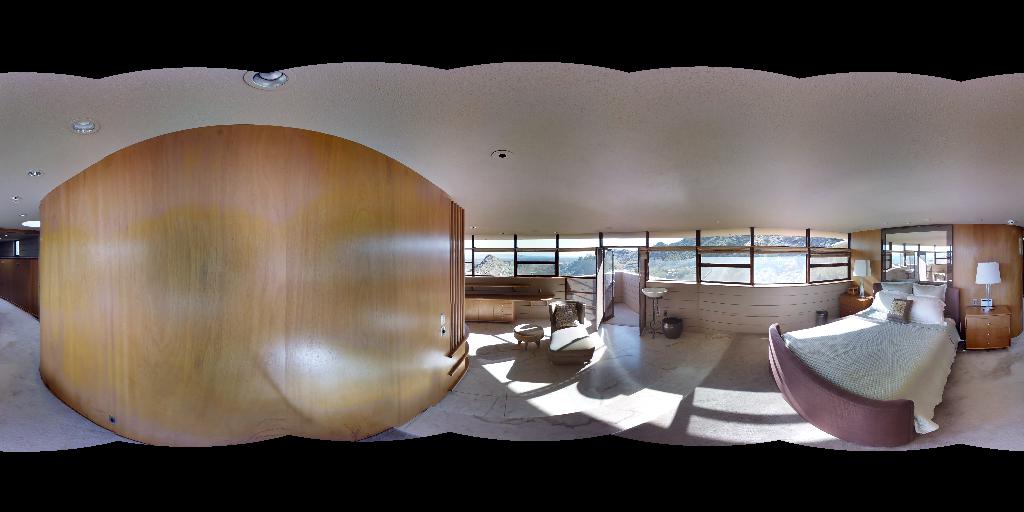} & \includegraphics[width=0.23\linewidth]{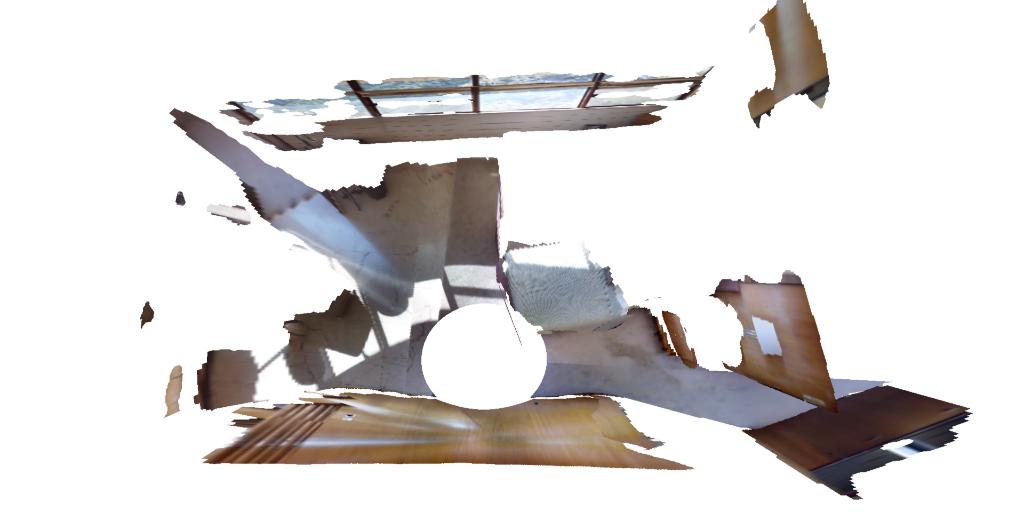} \\
        \includegraphics[width=0.23\linewidth]{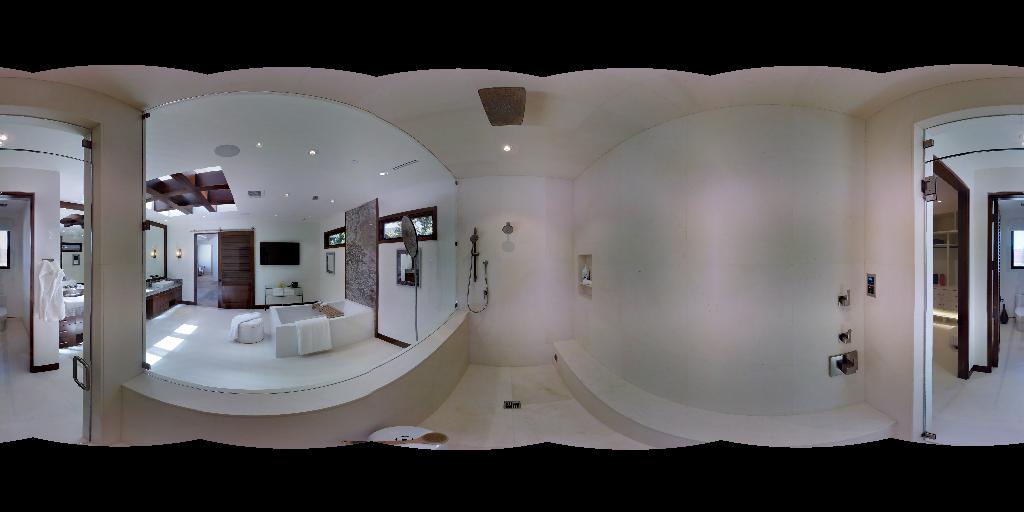} & \includegraphics[width=0.23\linewidth]{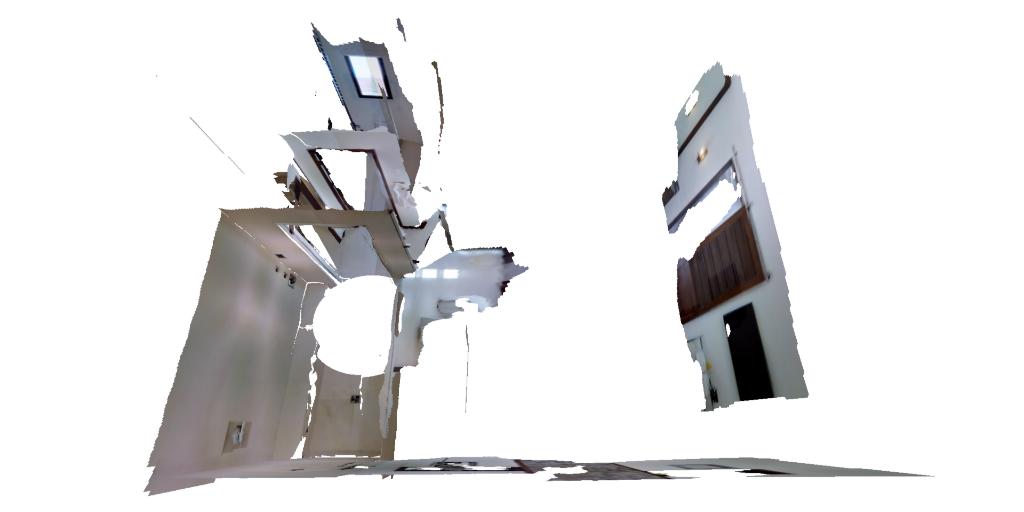} &
        \includegraphics[width=0.23\linewidth]{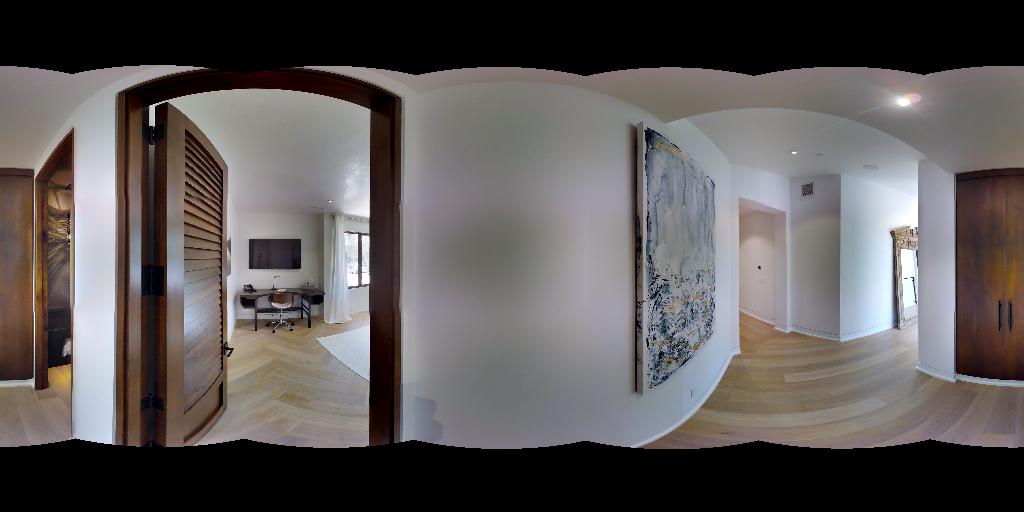} & \includegraphics[width=0.23\linewidth]{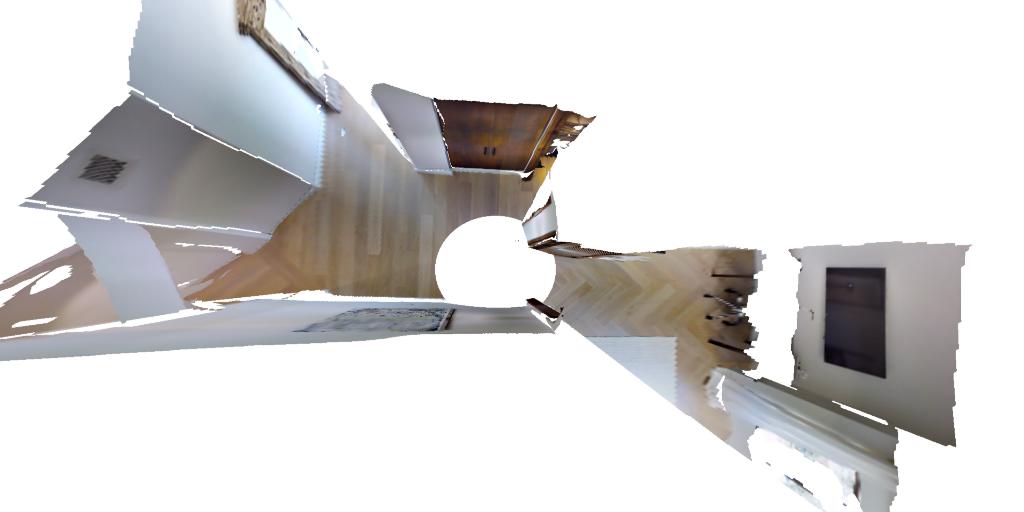} \\
        
        \includegraphics[width=0.23\linewidth]{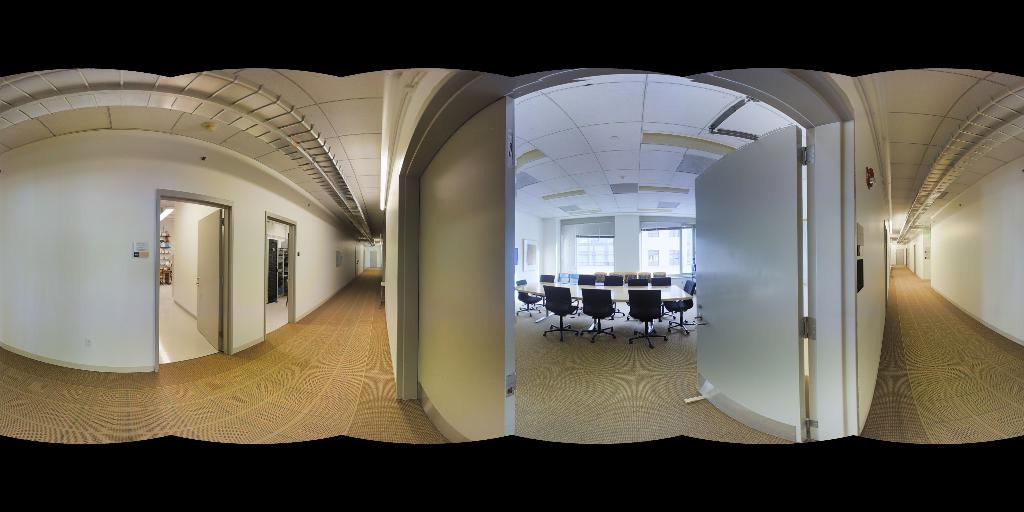} & \includegraphics[width=0.23\linewidth]{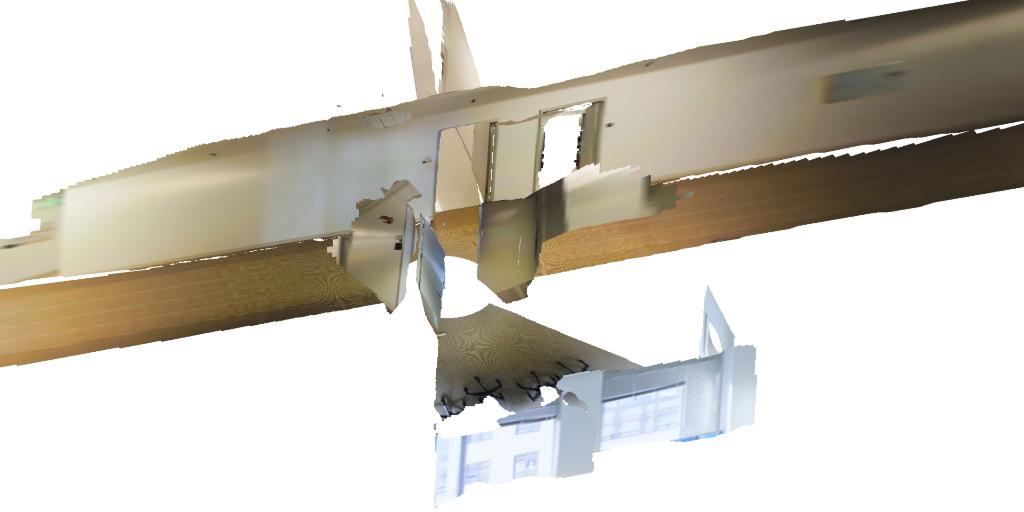} &
        \includegraphics[width=0.23\linewidth]{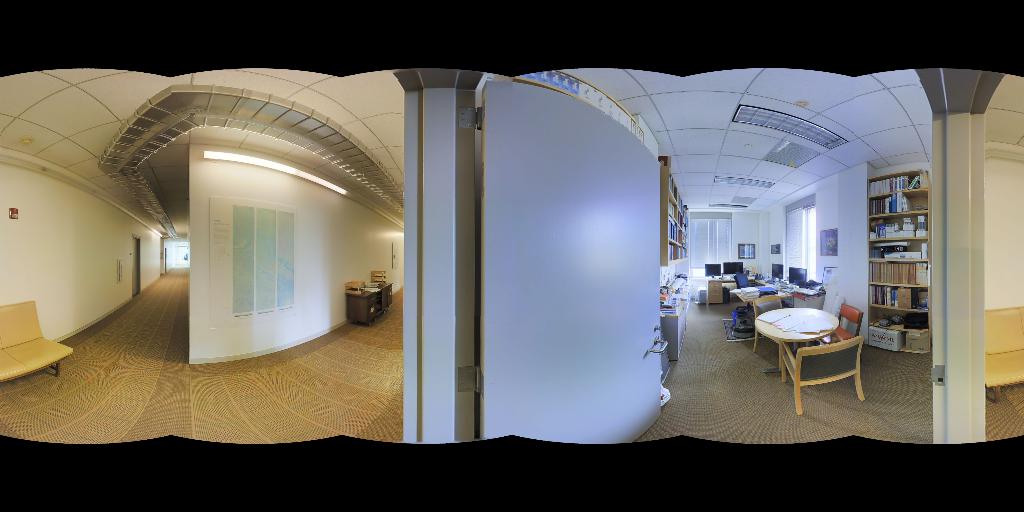} & \includegraphics[width=0.23\linewidth]{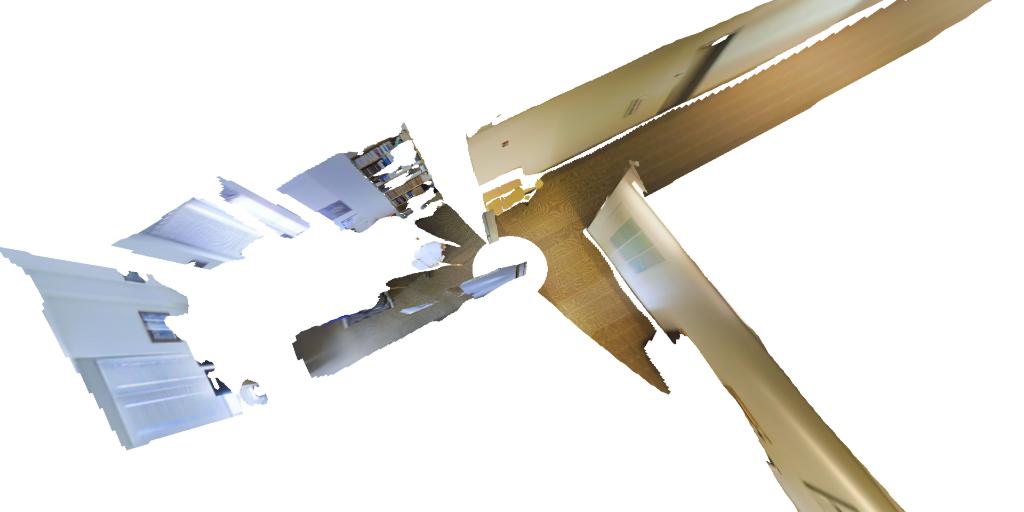} \\
        \includegraphics[width=0.23\linewidth]{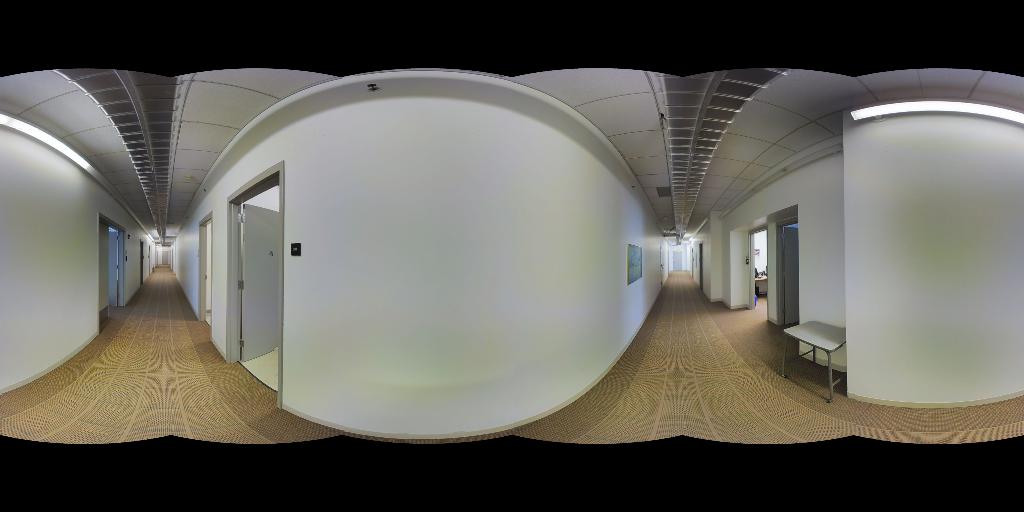} & \includegraphics[width=0.23\linewidth]{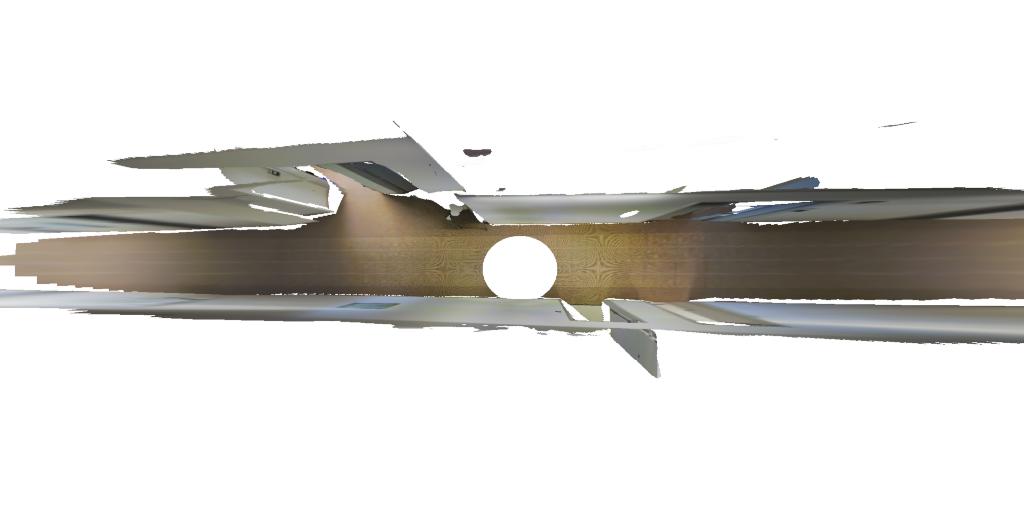} &
        \includegraphics[width=0.23\linewidth]{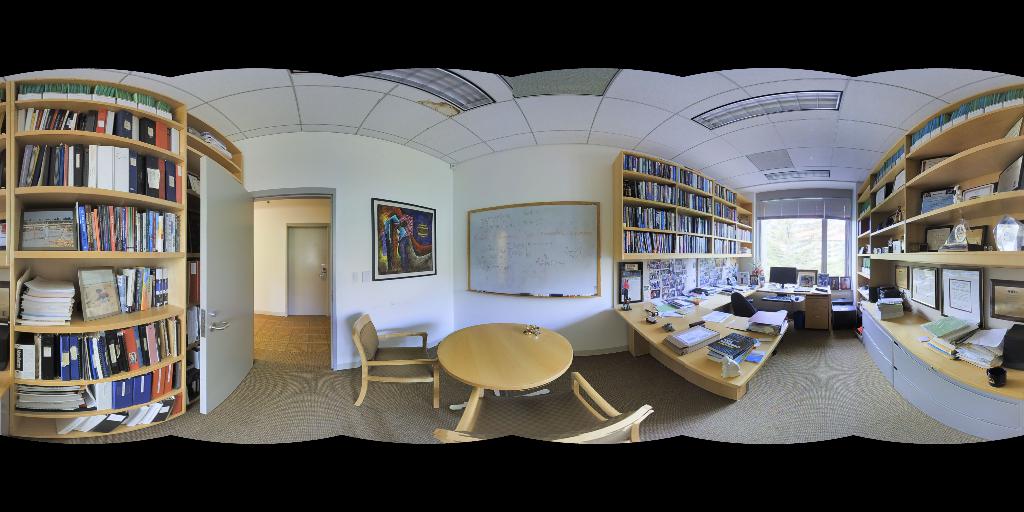} & \includegraphics[width=0.23\linewidth]{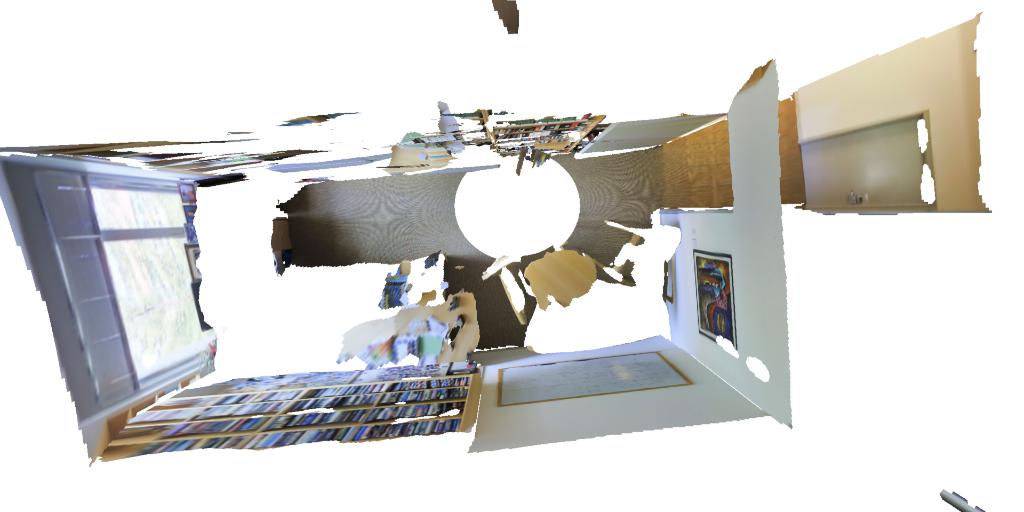} \\
        \includegraphics[width=0.23\linewidth]{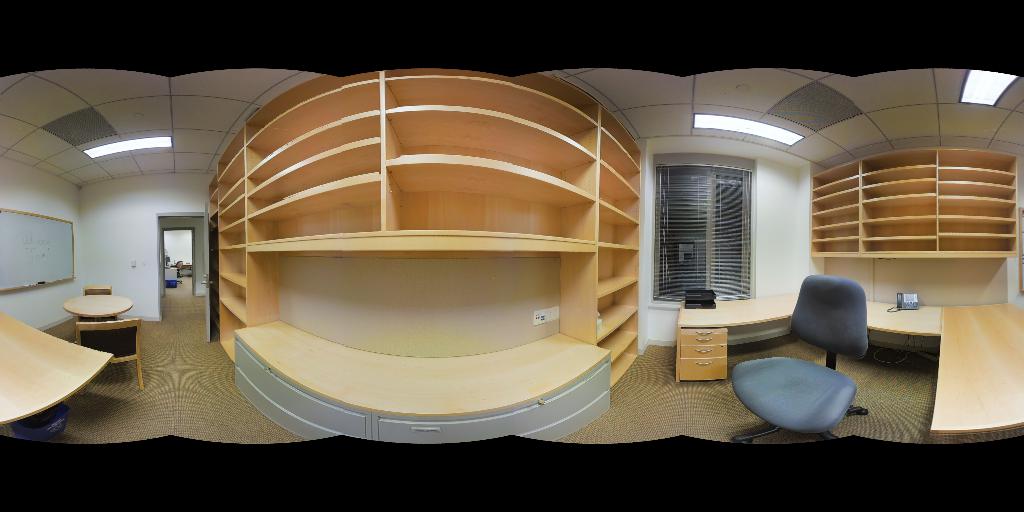} & \includegraphics[width=0.23\linewidth]{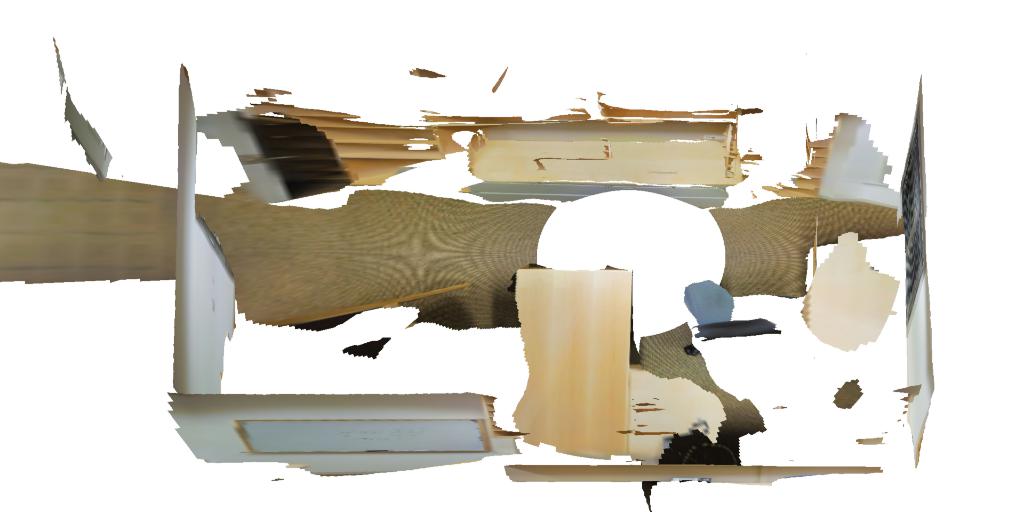} &
        \includegraphics[width=0.23\linewidth]{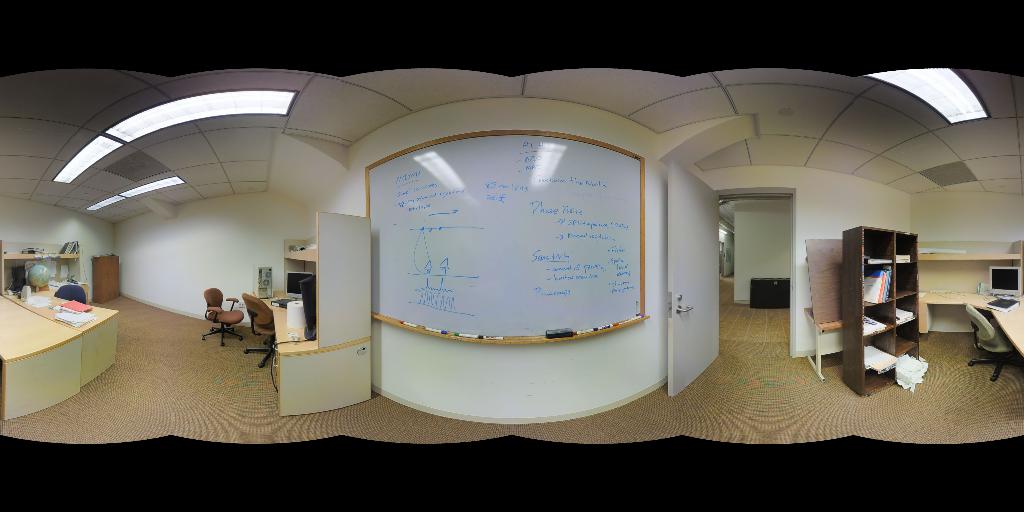} & \includegraphics[width=0.23\linewidth]{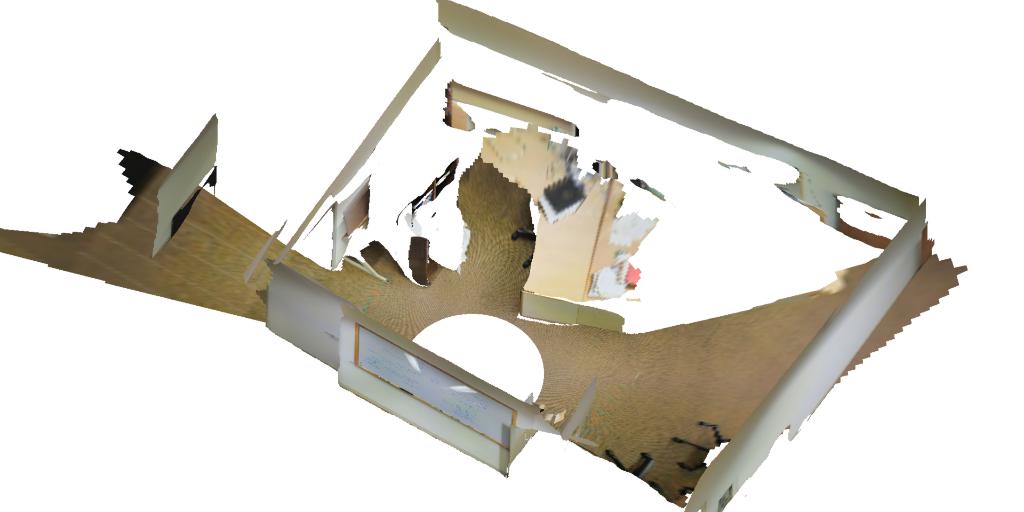} \\
        \includegraphics[width=0.23\linewidth]{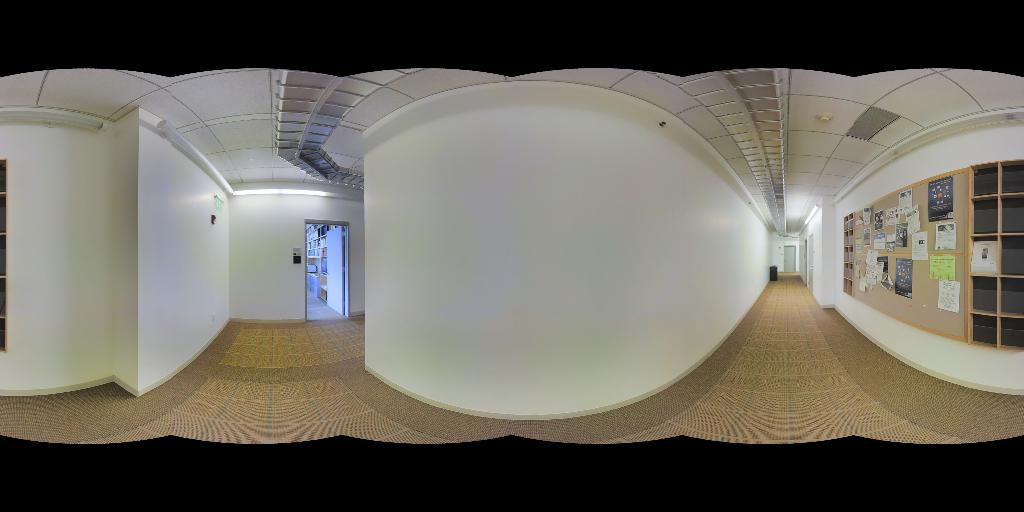} & \includegraphics[width=0.23\linewidth]{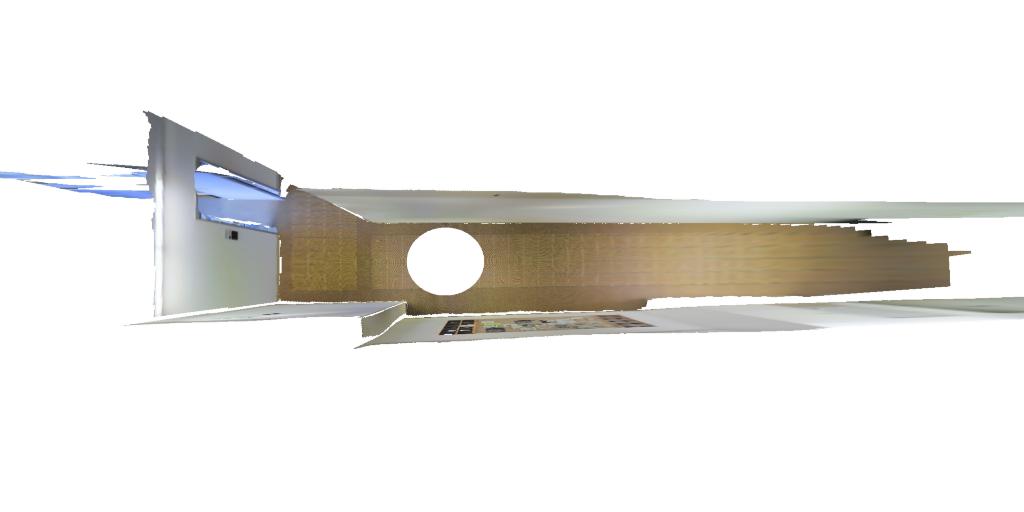} &
        \includegraphics[width=0.23\linewidth]{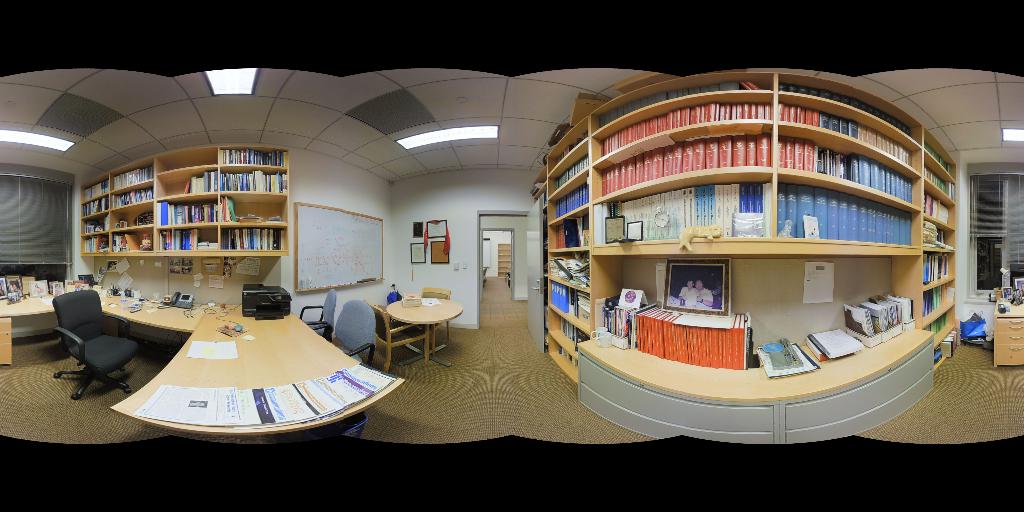} & \includegraphics[width=0.23\linewidth]{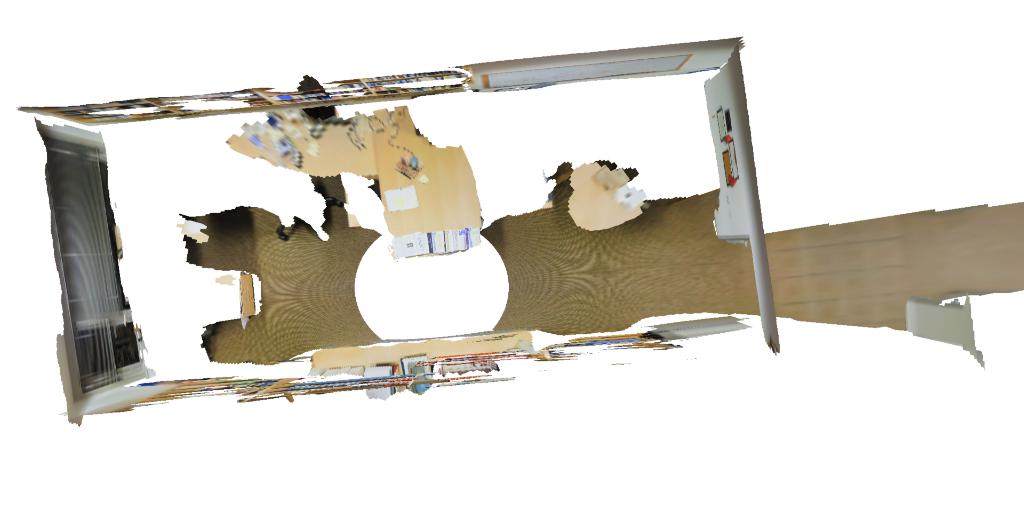} \\
        \includegraphics[width=0.23\linewidth]{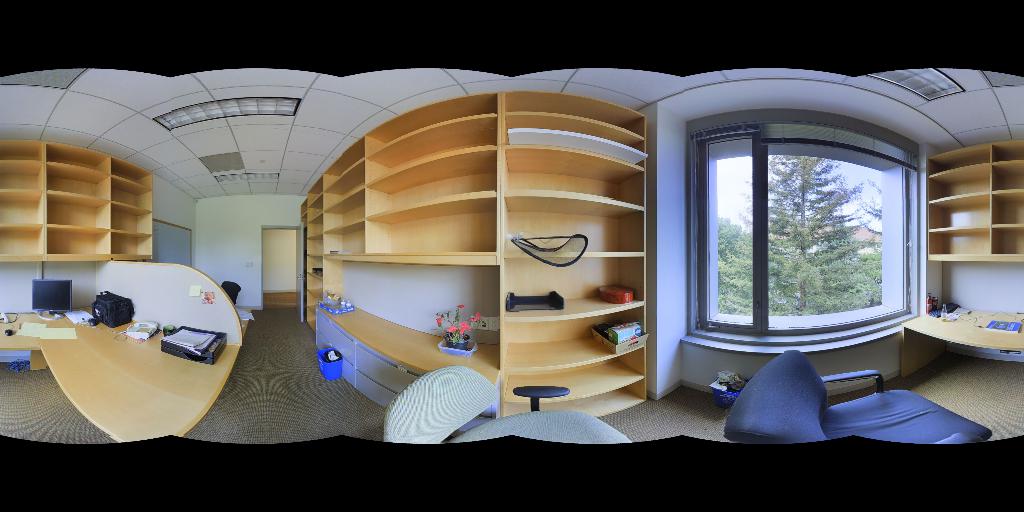} & \includegraphics[width=0.23\linewidth]{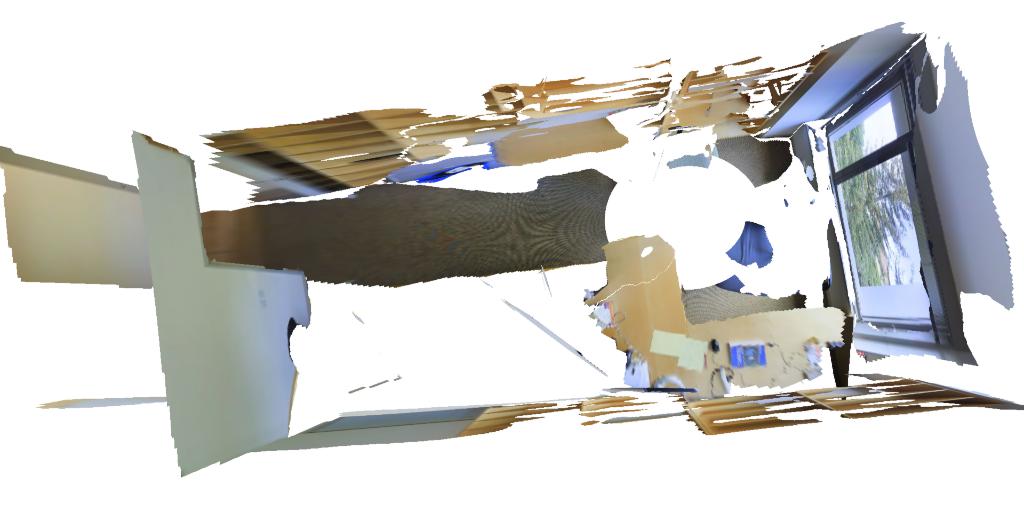} &
        \includegraphics[width=0.23\linewidth]{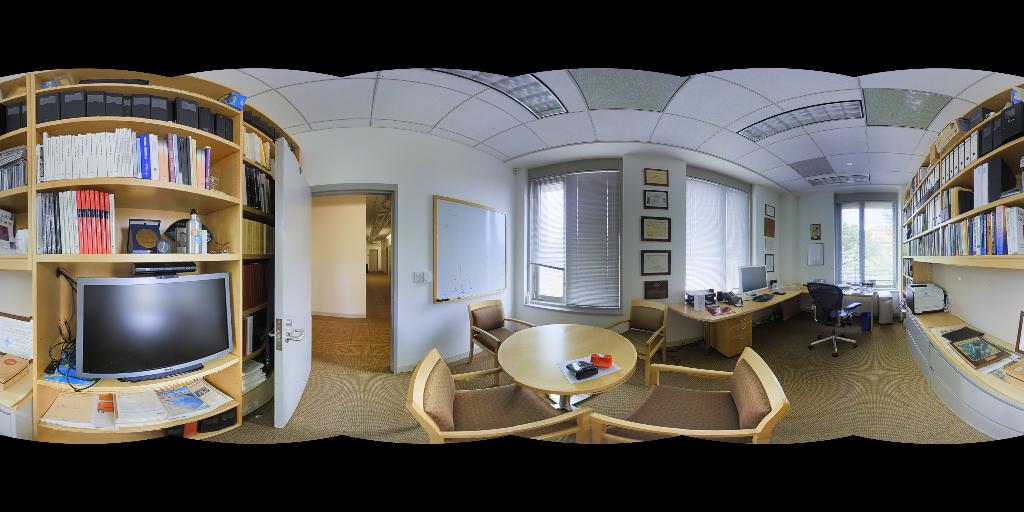} & \includegraphics[width=0.23\linewidth]{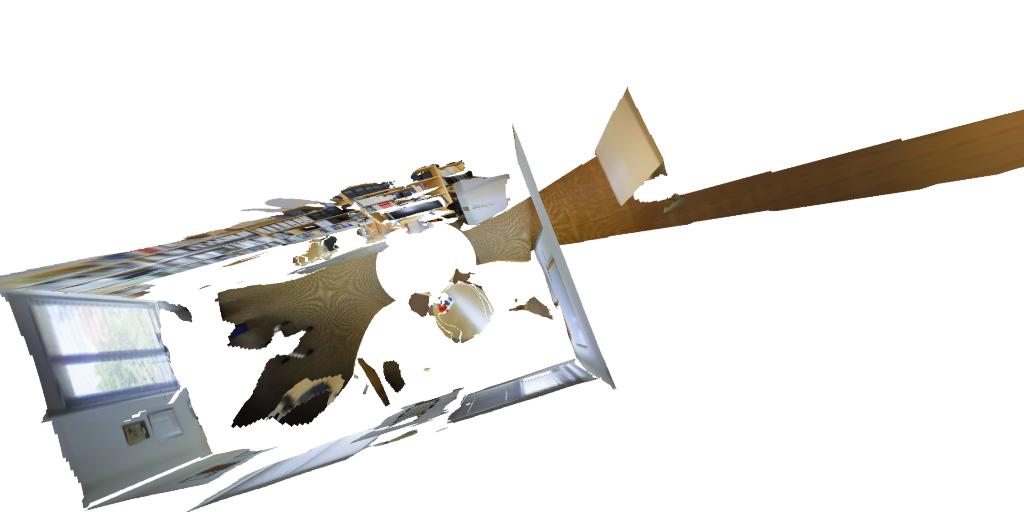} \\
    \end{tabular}
    \caption{
        3D mesh visualization from our approach.
    }
    \label{fig:more_qual_mesh}
\end{figure*}

\vspace{3em}
\noindent\textbf{Fig.~\ref{fig:more_quals_2d3d}~\ref{fig:more_quals_mp3d}~\ref{fig:more_quals_2d}~\ref{fig:more_qual_mesh} are in the next few pages.}

% \begin{figure*}
%     \centering
%     \includegraphics[width=0.99\linewidth]{fig/supp_2d_s2d3d.pdf}

%     \caption{
%         Qualitative comparison with baselines (See main text for detail)
%     }
%     \label{fig:more_quals_2d_s2d3d}
% \end{figure*}

% \begin{figure*}
%     \centering
%     \includegraphics[width=0.99\linewidth]{fig/supp_2d_mp3d.pdf}

%     \caption{
%         Qualitative comparison with baselines (See main text for detail)
%     }
%     \label{fig:more_quals_2d_mp3d}
% \end{figure*}

{\small
\bibliographystyle{ieee_fullname}
\bibliography{egbib}
}

\end{document}